%% file: SemivalueArbitraryGameable.tex
\documentclass[runningheads,11pt]{article}
\usepackage{fullpage}

\usepackage[utf8]{inputenc} 
\usepackage[T1]{fontenc}    

\usepackage{hyperref}

\usepackage{url}            
\usepackage{booktabs}       
\usepackage{amsfonts}       
\usepackage{nicefrac}       
\usepackage{microtype}      
\usepackage{xcolor}         
\usepackage{algorithm}
\usepackage{algorithmic}
\usepackage{graphicx}
\usepackage{subcaption}
\usepackage{microtype}
\usepackage{graphicx}
\usepackage{subcaption}
\usepackage{booktabs}
\usepackage{natbib}
\usepackage{graphicx}
\usepackage{amsmath}
\usepackage{amsthm}
\usepackage{enumitem}
\usepackage{cleveref}

\usepackage{enumitem}

\usepackage{algorithmic}
\usepackage{algorithm}
\usepackage{amsmath}
\newtheorem{definition}{Definition}

\newtheorem{proposition}{Proposition}
\newtheorem{assumption}{Assumption}
\newtheorem{lemma}{Lemma}
\usepackage{booktabs}
\usepackage{multirow, makecell}
\newcolumntype{L}[1]{>{\raggedright\let\newline\\\arraybackslash\hspace{0pt}}m{#1}}
\newcolumntype{C}[1]{>{\centering\let\newline\\\arraybackslash\hspace{0pt}}m{#1}}
\newcolumntype{R}[1]{>{\raggedleft\let\newline\\\arraybackslash\hspace{0pt}}m{#1}}
\usepackage{amsmath}

\usepackage{wrapfig}
\usepackage{caption}
\usepackage{subcaption}

\input{000_macros}

\def\balign#1\ealign{\begin{align}#1\end{align}}
\def\baligns#1\ealigns{\begin{align*}#1\end{align*}}
\def\balignat#1\ealign{\begin{alignat}#1\end{alignat}}
\def\balignats#1\ealigns{\begin{alignat*}#1\end{alignat*}}
\def\bitemize#1\eitemize{\begin{itemize}#1\end{itemize}}
\def\benumerate#1\eenumerate{\begin{enumerate}#1\end{enumerate}}

\newenvironment{talign*}
 {\csname align*\endcsname}
 {\endalign}
\newenvironment{talign}
 {\csname align\endcsname}
 {\endalign}
 \def\balignst#1\ealignst{\begin{talign*}#1\end{talign*}}
\def\balignt#1\ealignt{\begin{talign}#1\end{talign}}

\makeatletter
\def\@maketitle{%
  \newpage
  \begin{center}%
  \let \footnote \thanks
    {\LARGE \bf \@title \par}%
  \end{center}%
  \par}
\makeatother
\date{}
\title{Semivalue-based data valuation is  arbitrary and gameable} 

\begin{document}
\maketitle

\begin{center}
\vskip 1.5em
{\large
\begin{tabular}{ccc}
    \makecell{Hannah Diehl\\{\normalsize\texttt{hdiehl@mit.edu}}} & & \makecell{Ashia C. Wilson\\{\normalsize\texttt{ashia@mit.edu}}}
\end{tabular}
\vskip 0.25em
\normalsize
\begin{tabular}{c}
Department of Electrical Engineering and Computer Science, MIT
\end{tabular}}
\vskip 1.5em
\end{center}

\begin{abstract}
The game-theoretic notion of the \textit{semivalue} offers a popular framework for credit attribution and data valuation in machine learning. Semivalues have been proposed for a variety of high-stakes decisions involving data, such as determining contributor compensation, acquiring data from external sources, or filtering out low-value datapoints. In these applications, semivalues depend on the specification of a utility function that maps subsets of data to a scalar score. While it is broadly agreed that this utility function arises from a composition of a learning algorithm and a performance metric, its actual instantiation involves numerous subtle modeling choices. We argue that this underspecification leads to varying degrees of \emph{arbitrariness} in semivalue-based valuations. Small, but arguably reasonable changes to the utility function can induce substantial shifts in valuations across datapoints. Moreover, these valuation methodologies are also often \emph{gameable}:  low-cost adversarial strategies exist to exploit this ambiguity and systematically redistribute value among datapoints. Through theoretical constructions and empirical examples, we demonstrate that a bad-faith valuator can manipulate utility specifications to favor preferred datapoints, and that a good-faith valuator is left without principled guidance to justify any particular specification. These vulnerabilities raise ethical and epistemic concerns about the use of semivalues in several applications. We conclude by highlighting the burden of justification that semivalue-based approaches place on modelers and discuss important considerations for identifying appropriate uses. 
\end{abstract}

\section{Introduction}
\input{01_introduction}

\section{Related Work}
\input{02_related_work}

\section{Problem Formulation}
\input{03_problem_formulation}

\section{Experiments}\label{sec:experiments}
\input{04_experiments}
\section{Discussion}
\input{05_discussion}
\section{Acknowledgements}
We thank Alexander Tolbert for the useful comments. ACW acknowledges support from Simons Collaboration on Algorithmic Fairness. HD acknowledges the MIT EECS department and the Thriving Stars fellowship.

\newpage

\bibliographystyle{abbrvnat}
\bibliography{refs}

%%%%%%%%%%%%%%%%%%%%%%%%%%%%%%%%%%%%%%%%%%%%%%%
% APPENDIX
%%%%%%%%%%%%%%%%%%%%%%%%%%%%%%%%%%%%%%%%%%%%%%%

\newpage
\appendix
\onecolumn

\input{appendices/appendix_background}
\clearpage
\input{appendices/appendix_experimental}
\clearpage
\input{appendices/appendix_algorithms}
\clearpage
\input{appendices/appendix_proofs}
\end{document}

%% file: 000_macros.tex
\newcommand{\D}{\mathcal{D}}
\newcommand{\U}{\mathcal{U}}
\newcommand{\alg}{\mathcal{A}}

\newcommand{\aggF}{F_{\mathrm{agg}}(\psi(U);P)}

\newcommand{\indep}{\mathrel{\perp\!\!\!\perp}}

\newcommand{\testset}{\mathcal{T}}

%% file: 01_introduction.tex
Valuations are subjective. They are shaped by decisions about which properties are considered important. Even the market price, arguably the most concrete expression of value, reflects subjective assessments influenced by context, available information, and practical constraints. 
This inherent subjectivity prompts critical questions about how to value data, especially individual datapoints used to train machine learning models. 
As data plays an increasingly vital role in modern technological and economic systems, clarifying how its value should be determined grows more urgent. One popular approach 
employs methods from cooperative game theory, particularly semivalues such as the Shapley value, leave-one-out attribution. These methods are known to satisfy normative properties such as {\em fairness, symmetry,} and {\em efficiency,} which has motivated their adoption as principled approaches to data valuation. Within this formalism, individual datapoints act as players that agree on a utility function, which maps subsets of data to the performance of models trained on just those points. 
Semivalues, such as the Shapley value, then aggregate each datapoint’s marginal contributions across different subsets, providing an overall measure of its impact on downstream inference. %
To define utility in the context of machine learning, one typically composes a {\em learning algorithm}, which maps data to models, with a {\em model score} functional, such as loss, accuracy, or AUC. 

Notably, this is where substantial subjectivity reenters: there is no unique consensus mapping between learning tasks and utility functions. Model scoring evaluations are normatively chosen and contestable, and the specification of the learning algorithm, especially under data scarcity or ambiguous hyperparameter regimes, often lacks a standardized or even well-defined procedure. The utility functions defined within this underspecified design space can dramatically alter the inferred value of datapoints, and this ambiguity persists even when different choices yield equivalent full-data estimators with identical inferential value. This subtle yet consequential design space is frequently overlooked. While critiques of semivalue methods often emphasize their task-specificity, this focus obscures a deeper issue: even justifiable variations in utility definition can produce starkly divergent valuations, underscoring the need to treat utility specification itself as a central object of scrutiny.

Compounding this concern, we show that this sensitivity also opens the door to \emph{gaming}, easy-to-execute manipulation by a biased valuator. That is, simple algorithms can exploit the structure of semivalue computations to efficiently search the space of plausible utility specifications and deliver a utility function that favors a particular set of data contributors.
This vulnerability raises serious concerns about the reliability of these methods and fundamentally challenges the substantive fairness that motivates semivalue-based valuations in the first place.

Semivalues have been proposed as the basis for a range of consequential decisions, such as curating datasets~\citep{ghorbani_distributional_2020} and compensating data contributors~\citep{wang_principled_2020}. In applications such as dataset cleaning, value rankings are used to identify noisy or potentially corrupted observation to be removed from the dataset~\citep{namba_thresholding_2024}. In practice, many approaches rely on an unacknowledged assumption that the utility function is fully specified by the decisions made regarding learning algorithms and model evaluation in the normal course of the underlying machine learning task. In exemplary cases, such as~\cite{tang_medical_2021}, results are bolstered by 
repeating data valuation with alternative model architectures and test sets to show robustness of valuation outcomes. However, this step is often omitted, likely because even a single semivalue computation is computationally expensive. Furthermore, there are no established norms for comprehensively enumerating and evaluating all entailed choices. As a result, many decisions in utility specification go uninterrogated. This risks entrenching arbitrary or ill-suited choices, embedding subjective assumptions into systems under the guise of objectivity. When left unexplored, these defaults have the power to shape which datapoints are amplified or erased and whose contributions are credited or ignored without principled justification.

\paragraph{Contributions.} We introduce a unified framework for understanding and evaluating arbitrariness of semivalue‐based data valuations.

\noindent \textbf{(a) A formal model of valuation arbitrariness.}  
We show that every semivalue‐based data valuation relies on having a well-defined counter-factual model whose specification rests on a host of under‐specified choices: how to transform raw performance scores, weight false positives versus negatives, and handle tiny training coalitions. These choices do not affect the selected model but can dramatically reallocate credit among datapoints.  We capture this latitude by defining an “ambiguity set” of utility specifications that are observationally equivalent at the level of model‐selection, thereby making explicit the hidden degrees of freedom in any valuation pipeline. We illustrate this concept by highlighting three specific “ambiguity sets” that impact a wide range of analyses.

\noindent \textbf{(b) Quantitative measures of specification impact.}  
To assess the real‐world consequences of these arbitrary choices, we introduce a suite of group‐ and individual‐level metrics, such as normalized share of total value, rank quantile, and survival rate under low‐value filtering, that directly reveal how different utility functions advantage or disadvantage particular (subsets of) datapoints.

\noindent \textbf{(c) Gameability.}  
We identify the existence of relatively low-cost algorithms to find a utility within a set of candidate functions that optimizes outcome favorability for a selected group.  This both exposes how valuations can be adversarially manipulated and provides a tool for selecting specifications that safeguard against bias arising from epistemic uncertainty.

%% file: 02_related_work.tex
\textbf{Foundations of data valuation.}
A growing literature treats data as an economic asset, ranging from intangible business property~\citep{cheong_investment_2023,moody_walsh} and private property~\citep{kleinberg_value_2001,jurcys_ownership_2020} to a driver of machine learning value~\citep{sim_data_2022}.  Market-based pricing and trading mechanisms have been explored at the intersection of economics and CS~\citep{agarwal_marketplace_2019,tian_private_2023,zhang_survey_2024}.  In ML, cooperative game theory, most notably the Shapley value~\citep{shapley_value_1952}, provides a principled means to attribute each datapoint’s inferential contribution~\citep{ghorbani_shapley_2019,jia_shapley_2019}.  Subsequent semivalue variants and approximation schemes (e.g.\ Data Banzhaf, beta Shapley) have expanded this toolkit~\citep{kwon_beta_2022,wang_banzhaf_2023}.  Applications include royalty-sharing in content generation~\citep{wang_economic_2024}, revenue distribution in healthcare datasets~\citep{zhu_incentive_2019}, denoising of large datasets~\citep{tang_medical_2021}, and fair crediting in federated learning~\citep{kumar_federated_2022}.  For a broad survey, see Zhang et al.~\citep[Sec.~8.2]{zhang_survey_2024}.

\textbf{Arbitrariness in valuation.}
Though the Shapley axioms guarantee certain equity properties within a given game, they say little about the broader fairness of the valuation process~\citep{dwork_abstracting_2020,barocas_fairness_ml}.  Fairness claims require contextual grounding, since arbitrary implementation choices can systematically advantage or disadvantage groups~\citep{bower_fair_2017,dwork_composition_2019}.  Recent work interrogates the moral dimensions of arbitrariness in ML pipelines~\citep{black_multiplicity_2022,ganesh_arbitrariness_2025,creel_leviathan_2022}, arguing that the \emph{systematicity} of arbitrary decisions, not arbitrariness per se, drives ethical concerns.  Our paper situates semivalue-based attribution within this discourse by quantifying how specification latitude can produce unjustified disparities and how adversarial valuators can pattern these disparities and beneficient valuators can act to unpattern them. This paper builds on our previous work~\citep{diehl_arbitrary_2025} establishing the fairness concerns arising from the arbitrariness in utility specification in semivalue-based data valuation by providing a systematic framework for analyzing outcome sensitivity and adversarial gameability.

%% file: 03_problem_formulation.tex
Given a dataset \(\mathcal{D} = \{z_j\}_{j=1}^{N}\), the goal of data valuation is to assign a numeric value $\psi_j$ to each datapoint \(z_j\). A prominent approach frames this task as a cooperative game, treating datapoints as players who collaboratively influence a model’s performance. To formally construct such a game, we first define a utility function that maps subsets of datapoints to a numerical utility, representing their collective contribution.
Once the cooperative game \((\mathcal{D}, U)\) is defined, the challenge becomes determining how to allocate value among individual datapoints based on their contributions. One widely studied class of valuation methods is known as \emph{semivalues}{\color{black}, which uphold desireable fairness axioms (Appendix~\ref{app:further_background})}. Semivalues assign a datapoint’s value by averaging its incremental contributions across subsets of different sizes, weighted according to a specific distribution.
Formally:
\begin{definition}[Semivalue] Given a set of \(N\) datapoints (players) with cardinality \(n = |N|\), a utility function \(U\), and a set of non-negative weights \(\{w_k\}_{k = 0}^{n - 1}\) summing to one, the semivalue for each datapoint \(j\) is given by:
\[
 \psi_j(U, w) := \sum_{S \subseteq \D \setminus \{z_j\}} w_{|S|} \cdot \left[U(S \cup \{z_j\}) - U(S)\right].
\]
\end{definition}
Several prominent semivalues have been proposed in recent literature, including the Data Shapley~\citep{ghorbani_shapley_2019}, Data Banzhaf~\citep{wang_banzhaf_2023}, and the simpler leave-one-out (LOO) valuation. These methods differ primarily in how they select weights \( w_k \). The Shapley value assigns weights inversely proportional to the number of subsets of a given size,  with \( w_k = (n \cdot \binom{n - 1}{k})^{-1} \). The Banzhaf value places uniform weight across all subsets, with \( w_k = 2^{1-n} \), irrespective of subset size. In contrast, LOO places all its weight on the subset of size \( n - 1 \), effectively evaluating the direct incremental contribution of a datapoint when added to the full dataset. A detailed discussion of the axiomatic foundations underlying these choices is provided in Appendix~\ref{app:further_background}.

Semivalues rely on a counterfactual interpretation of utility being well-specified. In the context of games, this means that the utility assigned to each subset of players must have a clearly defined and consistent meaning across players. This condition ensures that the marginal contribution of an individual player can be meaningfully computed by comparing coalitions that differ only in that player’s inclusion.
In the context of machine learning, the prevailing formalization defines the utility function as the composition of two parts. The first is a statistical algorithm $\mathcal{A} : 2^{\mathcal{D}} \to \mathcal{F}$ that maps any subset of datapoints $S \subseteq \mathcal{D}$ to a trained model $f \in \mathcal{F}$. The second is a performance metric $V : \mathcal{F} \to \mathbb{R}$, which evaluates the quality or effectiveness of the model, often referred to as a \emph{model score}. {\color{black} The composition \(U = V \circ \alg\) thus offers a mapping from a collection of observations to a scalar-valued score, as required by the semivalue framing.}
This formalism assumes that the model and its score are stable and interpretable under changes to the dataset. That is, the utility of a coalition should reflect a meaningful and consistent counterfactual outcome. If this condition fails, e.g., due to retraining instability, randomness in $\mathcal{A}$, or sensitivity in $V$, then the counterfactual comparisons that semivalues rely on become ill-defined, compromising the resulting attributions.

\subsection{Quantifying outcome favorability}
\label{sec:quantifying}
To quantify how advantageous a utility specification $U \in \mathcal{U}$ is for a target subset $P \subseteq \mathcal{D}$, we introduce a \emph{favorability function}
$
F\colon (\psi, P) \mapsto \mathbb{R},
$
where $\psi_i(U)$ denotes the semivalue assigned to datapoint $i$ under utility function $U$. {\color{black} Intuitively, \(F\) maps semivalue outcomes for a dataset to a real-valued scalar corresponding to how favorable this outcome is to preferenced subset \(P\).}
This formalism systematizes the translation of semivalue assignments into practical outcomes, allowing us to compare how different utility specifications affect the valuation of individual datapoints or subgroups of data. In doing so, it reframes the abstract problem of utility selection in terms of its material consequences to contributing datapoint providers.
Based on the literature, we identify three examples of favorability measures based on common downstream use cases:

\begin{itemize}[leftmargin=*]
    \item \textbf{Aggregate value.} When downstream decisions are informed by the magnitudes of the semivalues~ \citep{wang_economic_2024, agarwal_marketplace_2019}, the favorability of an outcome for a group \(P\) will be based on the group's aggregate value, \(\aggF=\sum_{i\in P}\psi_i(U)\).
    \item \textbf{Rank.} As ML performance does not directly correspond to a concrete reward to distribution, many have reconsidered the efficiency axiom and the interpretation of semivalues in the absolute sense~\citep{kwon_beta_2022,wang_banzhaf_2023}. This leads to proposals to base decisions on an observation's rank within its dataset, e.g. top-n credit attribution. These applications motivate a rank favorability metric, \(F_{\mathrm{rank}}(\psi(U);\{i\})=\mathrm{rank}(i,\psi_i(U))\).\footnote{Summary statistics of ranks of individuals in a group have limited practical value. As such, we only apply this favorability metric to singleton \(P\). Furthermore, for integer index \(i\) and vector \(x\), we use \(\mathrm{rank}(i,x)\) to denote the rank (ascending) of \(x_i\) among components of \(x\), that is \(\mathrm{rank} (i,x)=\sum_{i'=1}^{|x|}\mathbf{1}(x_i>x_{i'})\).}
    \item \textbf{Filter survival.} A common data valuation benchmarking task is based on prevalent interest in the use of data values for removing `noisy' or corrupted observations from datasets~\citep{schoch_data_2023,enshaei_data_2021}. This inspires a favorability metric based on filter survival, such as the fraction of a subset that survives a filter of the bottom \(\alpha\)-fraction of the dataset, \(F_{\mathrm{filt}}(\psi(U);P)
    = \frac{1}{|P|}\sum_{i\in P}\mathbf{1}\bigl[\mathrm{rank}(i;\psi(U)) > \alpha\,|\mathcal{D}|\bigr]\).
\end{itemize}

\subsection{Utility Arbitrariness}\label{sec:arbitrariness}

Model utility cannot be observed directly; it rests on normative judgments about how to weight different model errors, which hypothetical objective the model optimized, or which counterfactual scenarios one deems relevant.  These choices are inherently subjective and often under-specified, introducing an unavoidable epistemic gap between inference and value.  We gather all utility functions that arguably inhabit this gap and quantify its impact using a favorability metric $F$ on any subset $P\subseteq\mathcal{D}$. We define the worst-case variation as
\begin{align}
\label{eq:range}
\text{range}(\mathcal{U};F) =\max_{U,U'\in\mathcal{U}}[F(\psi(U);P)-F(\psi(U');P)].
\end{align}
This range measures how much equally defensible mappings from inference to value can sway semivalue-based attributions. We now outline the three ambiguity families we consider.

\subsubsection{Algorithm Ambiguity}\label{sec:ambiguitity}
When quantifying an individual observation's contribution to the performance of the model output by learning algorithm \(\alg\) trained on \(\D\), 
the only behavior of \(\alg\) that is, in fact, strictly prescribed by the underlying learning task is \(\alg(\D)\). Though standard machine learning algorithms  have meaningfully defined behavior on other datasets of similar scale, the training procedure used for the full dataset \(\D\) may be ill-posed or undefined on datasets of a much smaller scale.
As general semivalue-based valuations aggregate marginal contributions over many coalitions \(S\subseteq\mathcal{D}\), including singletons and other small subsets, a careful analyst must consider how behavior of \(\alg\) is defined on  coalitions of all scales. The algorithm's behavior for the largest regime of cardinalities should coincide with the algorithm's behavior on the full dataset in the underlying machine learning task. However, an analyst could be justifed in introducing alternative behavior for limited-cardinality coalitions, as the `small-data' regime demands special consideration~\citep{bornschein_small_2020}. Modification to \(\alg\) to accomodate these coalitions may include increased regularization, restricted model complexity, modified approaches to model selection, or even, particularly for the smallest of coalitions, the option to return an untrained model.  In fact, code provided by the \verb|opendataval| package offered by~\citep{jiang_opendataval_2023} incorporates a subtle default behavior to ignore coalitions with fewer than 5 observations, independent of the learning algorithm or nature of the dataset. {\color{black} When this default behavior is overlooked or a value is set without rational basis for a novel problem, this becomes an arbitrary small data threshold.}
The requirement to specify behavior for small coalitions, and the need to select a size threshold that defines which coalitions are `small' in the context of a given task, introduces ambiguity into its specification.

In order to reason about the impact of this under-specification, we introduce the following notion of \emph{small-data algorithmic ambiguity} set associated with a learning algorithm \(\alg\) and performance metric \(V\).
\begin{definition}[Small-Data Algorithmic Ambiguity]
\label{def:small-data-ambiguity}
Let $k_{\min} \in \mathbb{N}$ define the small-data regime, and let $\mathcal{S}_{k_{\min}} = \{S \subset \mathcal{D} : |S| < k_{\min}\}$ denote the set of all subsets of $\mathcal{D}$ with fewer than $k_{\min}$ elements.
We say that two utility functions $U_0 = V \circ \alg_0$ and $U_1 = V \circ \alg_1$ are ambiguous under $k_{\min}$-small-data algorithmic ambiguity if:
(1) both are defined using the same performance metric $V$; and (2) the learning algorithms $\alg_0$ and $\alg_1$ produce identical outputs on all subsets outside the small-data regime.
The associated ambiguity set is defined as:
$$
\mathcal{U}_{\mathrm{small}}(k_{\min}) = \left\{ V \circ \alg' \mid \alg'(S) = \alg(S) \quad \forall S \subseteq \mathcal{D}, |S| \geq k_{\min} \right\}.
$$
This contains all utility functions that differ only in their behavior on small-data subsets.
\end{definition}

The notion of a `small-data regime’ is inherently context-dependent. In clinical trials, datasets with fewer than 50 samples may be informative; in contrast, large language models may require thousands of samples before yielding meaningful output. The choice of threshold $k_{\min}$ should reflect (1) the inductive limitations of the learning algorithm and specifically, the minimal dataset size for which the unmodified procedure is well-defined; and (2) the application-specific context in which data valuation is performed.
In practice, the selection of $k_{\min}$ and the construction of $\U_{\mathrm{small}}(k_{\min})$ should be guided by the analyst’s domain knowledge and understanding of the modeling pipeline.

For example,
suppose an analyst trains an ordinary least squares (OLS) regression model on 1000 records with 100 covariates. The fitted model is used to inform inventory decisions, and the resulting profit is allocated using Data Shapley. Notably, utility evaluation for subsets with fewer than 100 records yields singular design matrices, making OLS undefined. This forces the analyst to confront the ambiguity of how utility should be extended to these coalitions.
Several resolution strategies are possible. For example,
 the analyst may reason that, under a counterfactual with only 80 samples, regularization (e.g., ridge regression) would have been employed. This provides justification to redefine $\alg(S)$ as a regularized estimator when $|S| < 100$.
Alternatively, domain knowledge may suggest that subsets below a certain size are inherently uninformative, e.g., no algorithm could reliably guide purchasing decisions. In this case, the analyst might assign such coalitions the same utility as the empty set.
{\color{black} Selecting an appropriate approach for the small data regimes requires domain-informed assessments, such as evaluating the risks of generalization error or determining proper handling of extreme observations.}
These decisions reflect the structural and contextual limitations of the learning algorithm and introduce a spectrum of plausible utility functions within $\U_{\mathrm{small}}(k_{\min})$.

\begin{definition}[Untrained Small-Coalition Candidates]
\label{def:untrained-fallback}
Let \(k^* \in [0, |\mathcal{D}|)\) be the maximal `small-data' cardinality threshold. Define the augmented learner \(\mathcal{A}_{k}\) as the standard learner \(\mathcal{A}\) with a fallback returning an untrained model:
\(
\mathcal{A}_{k_{\min}}(S) = \{\alg(S) \text{ if } |S| \geq k_{\min}, \text{ else } \alg(\emptyset)\}.
\)
Then, the set of untrained small-coalition candidate utilities is:
\[
\mathcal{U}_{0}(k^*) = \{ V \circ \mathcal{A}_{k_{\min}} \mid k_{\min} \in [0, k^*] \},
\]
consisting of utilities that match a baseline utility \(V \circ \mathcal{A}\) on sufficiently large coalitions and return the fallback value \(V(\mathcal{A}(\emptyset))\) for coalitions below the threshold \(k_{\min}\).
\end{definition}

\subsubsection{Model Score Ambiguity}
The model score $V$ aims to reflect the utility of the trained model, ideally capturing its economic or societal value in deployment. However, since directly measuring utility is typically infeasible, empirical proxies are commonly employed. For regression tasks, a prevalent practice is to define $V$ using the training objective (e.g., negative loss), under the assumption that minimizing loss aligns with maximizing utility. Yet, this assumption leads to an inherent underdetermination: any strictly increasing transformation $f$ of the loss results in the same model selection and preserves the ordering of model scores. More broadly, in many evaluation contexts, decisions rely exclusively on model rankings rather than absolute scores. Because monotonic transformations preserve rankings, any such transformation of performance metrics yields indistinguishable evaluation outcomes. We formalize this inference ambiguity in the following definition: 

\begin{definition}[Score-Transformation Ambiguity]
\label{prop:score-transformation-ambiguity}
We say two utility functions $U_0 = V \circ \alg$ and $U_1 = V' \circ \alg$ are ambiguous under score-transformation ambiguity if (1) they are defined using the same learning algorithm $\alg$; and (2) the performance metrics $V$ and $V'$ are related by a strictly increasing transformation. The associated ambiguity set is:
$$
\mathcal{U}_{\mathrm{mono}} = \left\{ V' \circ \alg \;\middle|\; V'(\theta) = f(V(\theta)) \text{ for some strictly increasing } f \colon \mathbb{R} \to \mathbb{R} \right\}.
$$
This contains all utility functions that differ only by monotonic transformations of the score.
\end{definition}
For regression tasks, where training and evaluation often coincide, \(\mathcal{U}_{1}\) provides a natural ambiguity set even if in practice we evaluate only a small discrete subset (see Sec.~\ref{sec:experiments}).  
In classification, however, models are trained with a probability‐based surrogate (e.g.\ cross‐entropy) but evaluated on thresholded decisions (accuracy, F1, AUC, etc.), each implicitly encoding a particular false‐positive vs.\ false‐negative trade‐off. Since the true error costs are seldom agreed upon, practitioners default to convenient settings that obscure substantial epistemic uncertainty.
An appropriate performance function may incorporate a revenue model that attempts to estimate the actual monetary value of a model by considering the costs and profits from the downstream decisions they inform~\citep{agarwal_marketplace_2019}. The net-benefit metric for classification problems is an example of such a parameterized performance metric. 

\begin{definition}[Cost-Ratio Ambiguity]
\label{def:cost-ambiguity}
Let $T(S)$ and $F(S)$ denote the true and false positive rates of the classifier trained on coalition $S$, respectively.
Fix a cost-ratio interval $[a, b] \subset (0,1)$, and define the net benefit at cost ratio $p_t \in [a, b]$ as:
$
\mathrm{NB}(p_t; S) := T(S) - {p_t(1 - p_t)^{-1}}  F(S).
$
We say two utility functions are ambiguous under cost-ratio ambiguity if they are induced by the same classifier but differ only in the choice of $p_t \in [a, b]$.
The associated ambiguity set is:
$$
\mathcal{U}_{\mathrm{cost}}(a, b) = \left\{ \mathrm{NB}(p_t; \cdot) \;\middle|\; p_t \in [a, b] \right\}.
$$
This captures utility functions that vary due to differing relative costs of false and true positives.
\end{definition}
{\color{black} When decisions are made on the basis of estimated costs, such as the use of medical diagnostics to inform treatment decisions, uncertainty in the costs of false positives and false negatives introduces ambiguity in how models should be valued, which subsequently impacts data valuation outcomes.}
The use of revenue modeling and a net-benefit performance metric offers promise for principled reward distribution by faithfully mirroring the game-theoretic notion of utility: the actual reward that a coalition could obtain by `playing the game' alone. However, uncertainty in all stages of modeling costs and revenues may translate to utility specification choices that an analyst must navigate.

\subsection{Utility Gameability}\label{sec:gameable}
We say a valuation pipeline is gameable if an adversarial data valuator can feasibly identify the utilities that most favor a given group according to the favorability functions defined in~\ref{sec:quantifying}. We are interested in `feasibility' in comparison to the runtime needed to compute semivalues for a single utility. For the scope of this paper, we consider semivalues that are computed either (a) exactly, by scoring all coalitions, or (b) approximated by sampling. Valuation problems with  analytical solutions (eg KNN-shapley~\citep{jia_knn_2019}) are excluded from present consideration. We illustrate gameability in the simplest scenarios, and then move to more complex scenarios in which the structure of semivalues is leveraged for efficient gaming.

\begin{definition}[Exact Gameability]
\label{def:exact-gameability}
A candidate utility class $\mathcal{U}$ is said to be exactly gameable under favorability function $F$ and semivalue $\psi$ if, for any subset $P \subseteq \mathcal{D}$, there exists an algorithm that computes
$
U^* \in \arg\max_{U \in \mathcal{U}} F(\psi(U); P)
$
using at most $O(\mathrm{poly}(N))$ additional utility function evaluations beyond those required to compute $\psi(U)$ for each $U \in \mathcal{U}$.
\end{definition}
Given the baseline cost of computing most semivalues is $O(2^N)$, only modest overhead is needed to return an optimal utility in comparison. 
When exact semivalues are intractable, we define gameability relative to an approximate method.
\begin{definition}[$(\epsilon, \delta)$-Gameability]
\label{def:approx-gameability}
A candidate utility class $\mathcal{U}$ is $(\epsilon, \delta)$-gameable under favorability function $F$ and semivalue $\psi$ if there exists an algorithm that, for any subset $P \subseteq \mathcal{D}$, returns $U \in \mathcal{U}$ such that with probability at least $1 - \delta$,
$
\left| F(\psi(U), P) - \max_{U' \in \mathcal{U}} F(\psi(U'), P) \right| < \epsilon,
$
using at most $O( \mathrm{poly}(N) \log(1/\delta)/\epsilon^2 )$ utility function evaluations.
\end{definition}
\noindent  In order to facilitate runtime analysis of algorithms that involve semivalue approximation for datasets of variable size, we restrict present consideration to sets of utility functions with finite ranges. 
\begin{assumption} For dataset \(\D\) and utility candidate set \(\U\), there exists finite \(r\) such that, \[\max_{S\subseteq\D}[U(S)]-\min_{S\subset\D}[U(S)]\leq r, \quad \forall S\subseteq \D, U\in \mathcal{U}.\]
\end{assumption}
\noindent As argued in~\cite{wu_variance_2023}, many performance metrics relevant to machine learning applications, such as classification accuracy or recall, have fixed ranges by definition. In many other cases, the performance is upper bounded by definition, e.g. by squaring of errors, and a lower bound can be imposed by clipping values below some threshold, e.g. some fixed multiple worse than a naive model.

\textbf{Gameability of finite utility sets.} One of the simplest and hardest to detect ways an adversarial data valuator might manipulate valuations is by selecting from a fixed set of justifiable utility functions.
To demonstrate the construction of gameability results, we consider the example of optimization of \(F_{\mathrm{agg}}\) over these fixed size candidate sets.
\begin{proposition}[$(\epsilon, \delta)$-Gameability of Fixed Cardinality \(\mathcal{U}\)]\label{prop:fixed-gameability}
Let \(\mathcal{U}\) be a finite candidate set of utility functions of size \(O(1)\), i.e., independent of the dataset size \( |\mathcal{D}| = N\).
\(\mathcal{U}\) is $(\epsilon,\delta)$-gameable under favorability \(\aggF\) for any \(P\) using Algorithm~\ref{alg:discrete_search}.
\end{proposition}

The result above captures the setting in which a valuator enumerates a small set \(\mathcal{U}\) of justifiable utility functions and seeks to maximize the total value to a target subset \(P\). Algorithm~\ref{alg:discrete_search} performs an exhaustive evaluation of semivalue outcomes for each candidate, evaluates these outcomes under the favorability metric, and returns the optimizing utility. In settings where exact semivalue computation is tractable (e.g., via full coalition enumeration), this process yields \textit{exact} gameability.
In cases where semivalues are estimated via sampling, approximate gameability is possible as the probability that a suboptimal utility function is incorrectly identified as optimal decays rapidly with its deviation from the true optimum. See Appendix~\ref{app:approx_game} for a full probabilistic analysis.

In ambiguity classes defined by cardinality-dependent behavior, as may be encountered within \(\U_{\mathrm{small}}\) (Def ~\ref{def:small-data-ambiguity}), the size of the candidate may scale exponentially with dataset size. In such cases, enumerating over utility candidates (as in Algorithm~\ref{alg:discrete_search}) lacks \(\mathrm{poly}(N)\) complexity sought per~\ref{def:approx-gameability}. However, the ability to decompose all semivalues into strata according to into weighted sums of marginal contributions over fixed-size strata can be used to design feasible gaming algorithms. Algorithm~\ref{alg:min_k_search} shows gameability over \(\mathcal{U}_{0}(k^*)\) and Algorithm~\ref{alg:k_behavior_optimization} shows gameability over a subset of \(\U_{\mathrm{small}}\) when there are finite options for algorithmic behavior for each cardinality.

\begin{proposition}[$(\epsilon, \delta)$-Gameability of Small-Cardinality Behavior]\label{prop:small-gameability}
Let \(\mathcal{U}\subseteq \mathcal{U}_{\mathrm{small}}\) be the set of utility functions within the small-cardinality ambiguity set such that the behavior of the learning algorithm can be specified at for each cardinality by a constant number of possible behaviors. This candidate set is $(\epsilon,\delta)$-gameable under favorability  \(\aggF\) for any \(P\) using Algorithm~\ref{alg:k_behavior_optimization}.
\end{proposition}

\textbf{Gameability over infinite sets}
When \(\mathcal{U}\) is not discrete, it is not possible to simply approximate semivalue outcomes under all candidate utilities. An adversarial evaluator may choose a finite subset and attempt to find the optimal outcome over this subset of candidates, however this will not provide any near-optimality gaurantees in the general case. Still, some infinite cardinality candidate sets may permit gameability algorithms when they have structure that can be exploited. For example, the linearity of the net-benefit performance metric can be exploited to game aggregate values. %

\begin{proposition}[$(\epsilon,\delta)$-Gameability of $\U_{cost}$]\label{prop:cost-gameability}
Let \(\mathcal{U}_{cost}(a,b)\) be a utility ambiguity set arising from Definition~\ref{def:cost-ambiguity}. \(\mathcal{U}_{cost}\) is $(\epsilon,\delta)$-gameable under favorability \(F_{\mathrm{agg}}(\psi(U);P)\) for any \(P\) with respect to  \(\psi\) using Algorithm~\ref{alg:net_benefit_game} .
\end{proposition}

\subsection{Interaction with semivalue selection}

Due to differing weights on contributions from each cardinality of coalition, modifying the utility function specification will impact outcomes differently depending on the particular semivalue used. For example, Data Banzhaf places lower weight on small cardinalities compared to Data Shapley, reducing its sensitivity to challenges with specifying small coalition  behaviors. Nonetheless, Data Banzhaf remains arbitrary and gameable.

The Leave-One-Out (LOO) notion has unique structure that offer robustness to several types of underspecification. Though the LOO value has notable shortcomings for data valuation applications, prominently its lack of replication robustness~\citep{ghorbani_shapley_2019, han_replication-robust_2023}, this property may be advantageous in applications where LOO is appropriate. Specifically, LOO is invariant to ambiguity in algorithm behavior on all coalitions except those of size $N$ and $N{-}1$, since it depends only on the full dataset and leave-one-out subsets.
This simplicity confers robustness to underspecified modeling choices, including ambiguity over small-coalition behavior. Moreover, LOO is rank-stable (though not value-stable) under any strictly increasing transformation of the performance metric: for all strictly increasing $f$, the ranking induced by $f(V)$ matches that of $V$.
Although LOO remains sensitive to the choice of performance metric (e.g., evaluation metric, test set, or threshold), such sensitivity is intrinsic to the task and unavoidable. Furthermore, LOO admits exact computation with only $O(N)$ utility evaluations, offering a significant computational advantage over Shapley and Banzhaf, which require $O(2^N)$. This computational efficiency raises some concern for efficient exact gameability, but also enables more comprehensive robustness checks from good-faith evaluators.

%% file: 04_experiments.tex
\newcommand{\ptrange}{[0.5,0.6]}
We evaluate the gameability of semivalue-based data valuations on several real-world datasets identified as benchmarks for data valuation~\citep{jiang_opendataval_2023}.
Open-source datasets centering medical and social outcomes were prioritized during selection, as these represent applications in which arbitrariness and adversarial actions raise the most significant ethics concerns~\citep{fabris_algorithmic_2022}. Further details on selected datasets and baseline learning tasks can be found in Appendix \ref{app:exp_details}. 

Experiments explore the sensitivity of value- and rank-based outcomes to utility specification for three classification tasks and two regression tasks. Candidate utility sets were designed to concretely illustrate different types of underspecification explored in Section \ref{sec:arbitrariness} by isolating a single design choice to perturb. We consider the following candidate sets:
  \(\mathcal{U}_0\): Ambiguous utilities under small-coalition fallback, where all coalitions below size \(0.1|\mathcal{D}|\) map to \(U(\emptyset)\) (Def.~\ref{def:untrained-fallback});
\(\mathcal{U}_1\): Regression utilities from fixed algorithms and standard metrics \(V = \{\text{MSE}, \text{RMSE}\}\); a subset of \(\mathcal{U}_{\text{mono}}\) (Def.~\ref{prop:score-transformation-ambiguity}); and 
 \(\mathcal{U}_2\): Classification utilities based on cost ratios with threshold \(p_t \in \ptrange\), discretized into 100 utilities (Def.~\ref{def:cost-ambiguity}).
For all utility functions, semivalue outcomes for each observation in each dataset were computed via stratified sampling (Algorithm \ref{alg:sampler}).

We consider two favorability functions adapted to facilitate comparison across variable data sizes:
\vspace{-.4em}
\begin{align*}
 F_{\mathrm{payout}}(\psi(U);P) := \aggF \cdot {\tfrac{|\D|}{\sum_{j\in\D}\psi_j(U)}}  \text{ and }
F_{\mathrm{scaled~rank}}(\psi(U);P) := \tfrac{F_{\mathrm{rank}}(\psi(U);P)}{|\mathcal{D}|}.
\end{align*}

We compute \(\text{range}(F_P,\mathcal{U})\) for each of these favorability metrics for all singleton $P=\{i\}$ and 100 randomly sampled \(P\subseteq\mathcal{D}\) containing 10\% of \(\mathcal{D}\),  illustrating the extreme-case shifts an adversarial valuator could induce for arbitrary individuals or groups selected for preferential treatment.

\subsection{Results: Value Arbitrariness and Gameability}

\begin{figure}
    \centering
    \begin{subfigure}[t]{0.31\linewidth}
        \includegraphics[width=1.1\linewidth]{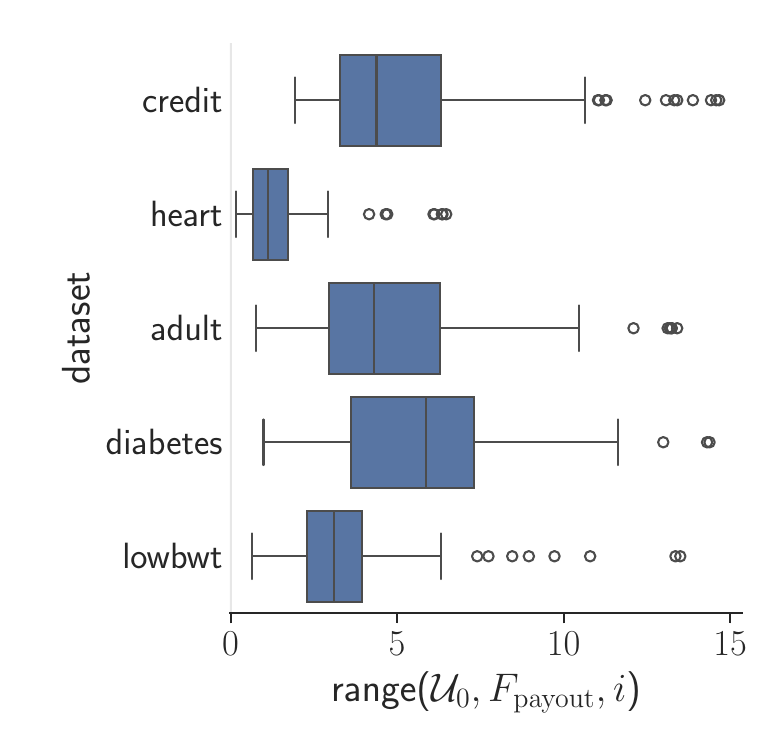}
        \caption{Under small coalition handling}
    \end{subfigure}
    \hfill
    \begin{subfigure}[t]{0.31\linewidth}
        \includegraphics[width=1.1\linewidth]{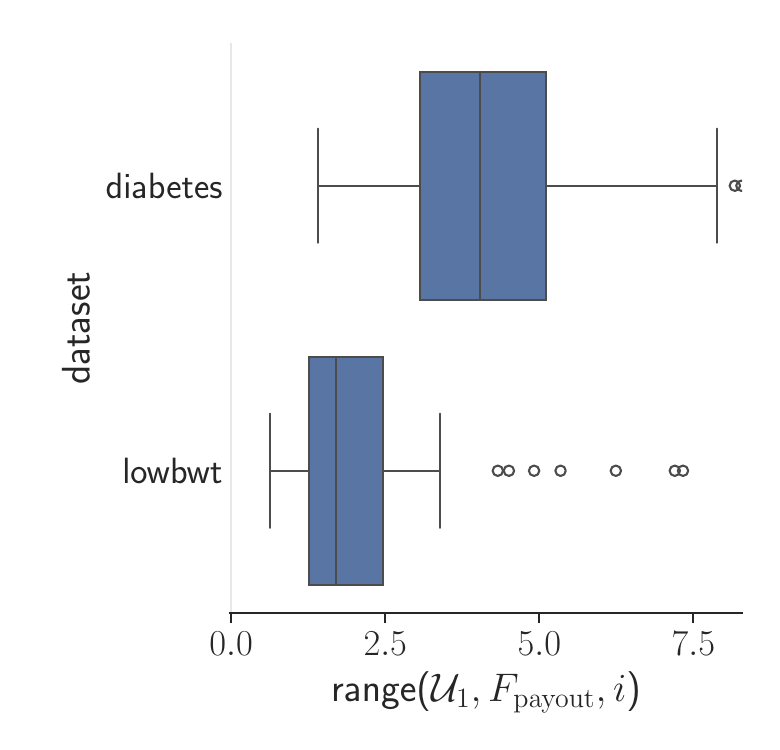}
        \caption{Monotonic \(V\) transformation}
    \end{subfigure}
    \hfill
    \begin{subfigure}[t]{0.31\linewidth}
        \includegraphics[width=1.1\linewidth]{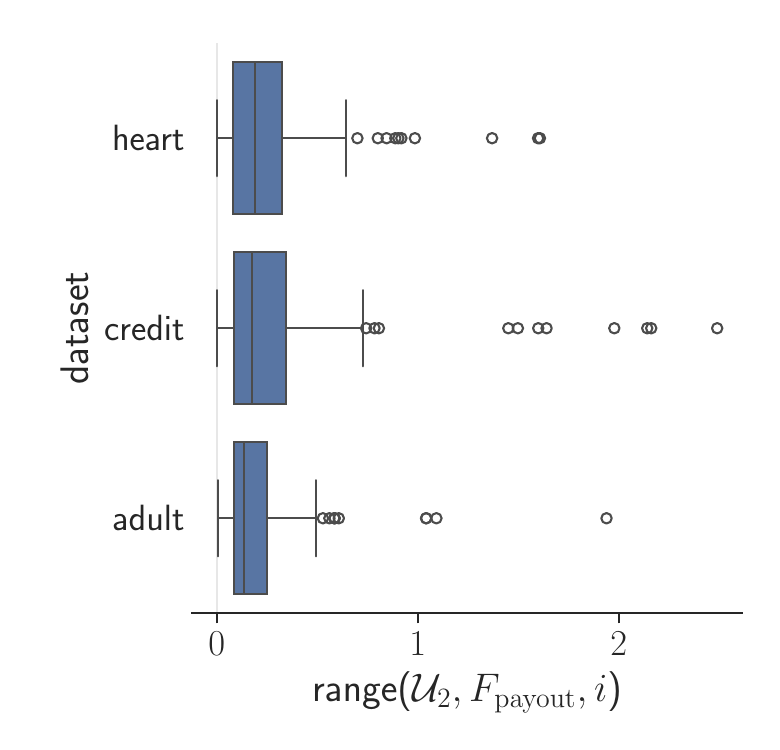}
        \caption{Cost-Ratio Parameter}
    \end{subfigure}
    \caption{Distribution of individual observations' payout sensitivity to each explored candidate alternative set (Sec.~\ref{sec:experiments})}
    \label{fig:vt_ind}
\end{figure}

Figure~\ref{fig:vt_ind} illustrates the sensitivity of each individual’s payout under a Data Shapley valuation. Each facet shows how an individual’s assigned value varies across candidate utility functions drawn from a single-choice ambiguity set. We focus on the Shapley value when analyzing absolute shifts in value, as its unique axiomatic property, that payouts sum to the total utility of the coalition, endows it with distinctive interpretive significance.

\( F_{\mathrm{payout}}\) normalizes the total reward budget such that the average individual payout is one cost unit. As such, differences in aggregate value favorability reflect structural variation in value allocation rather than dataset size or budget scale. With this metric 
\(\mathrm{range}(\mathcal{U}, F_{\mathrm{payout}}, i)\)
measures how much additional value individual $i$ can gain, relative to the least favorable utility, in units of average payout. For instance, $\mathrm{range} = 4$ implies $i$ can gain as much as four average contributors; a value near zero indicates robustness of the  Shapley value to utility choice within this candidate utility set \(\U\).

Results show significant payout sensitivity for the majority of points across all datasets. For example, in the credit dataset, the median individual's payout can be increased by about 5 units (with some up to 15) when switching between the worst and best small-coalition fallbacks. 
While some individuals have small value ranges under a particular candidate set, implying robustness of their shapley value to specification, most individuals have high valuation sensitivity to the same utility specification choice.

Analysis of the range of group payouts for randomized groups shows that an adversarial evaluator has the ability to favor demographic groups in aggregate without the gains of some individuals being offset by losses of others in a group. The corresponding visualization for group payouts can be found in Appendix~\ref{app:exp_details}.

\subsection{Results: Rank Arbitrariness and Gameability}

\begin{figure}
    \centering
    \begin{subfigure}[t]{0.31\linewidth}
        \includegraphics[width=1.1\linewidth]{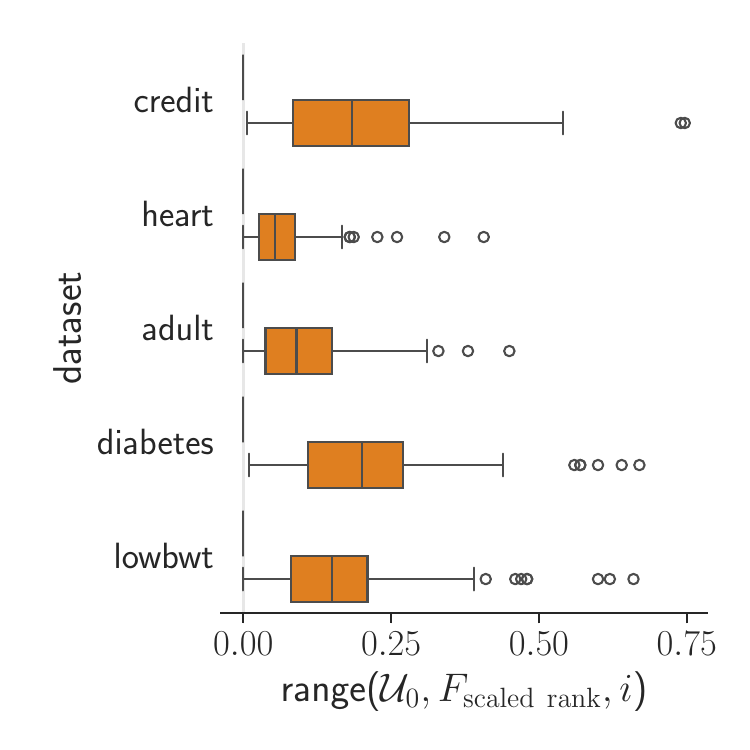}
        \caption{Under small coalition handling}
    \end{subfigure}
    \hfill
    \begin{subfigure}[t]{0.31\linewidth}
        \includegraphics[width=1.1\linewidth]{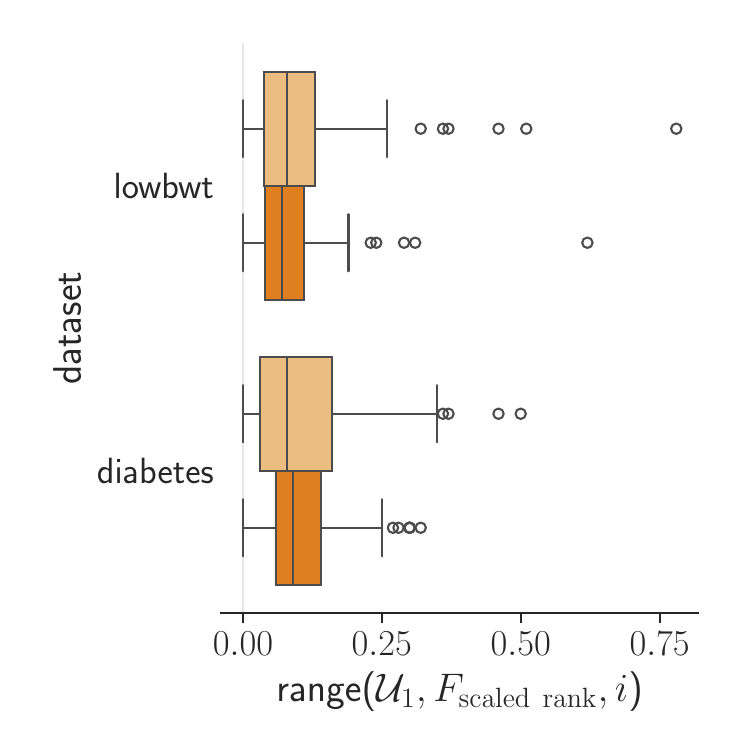}
        \caption{Monotonic \(V\) transformation}
    \end{subfigure}
    \hfill
    \begin{subfigure}[t]{0.31\linewidth}
        \includegraphics[width=1.1\linewidth]{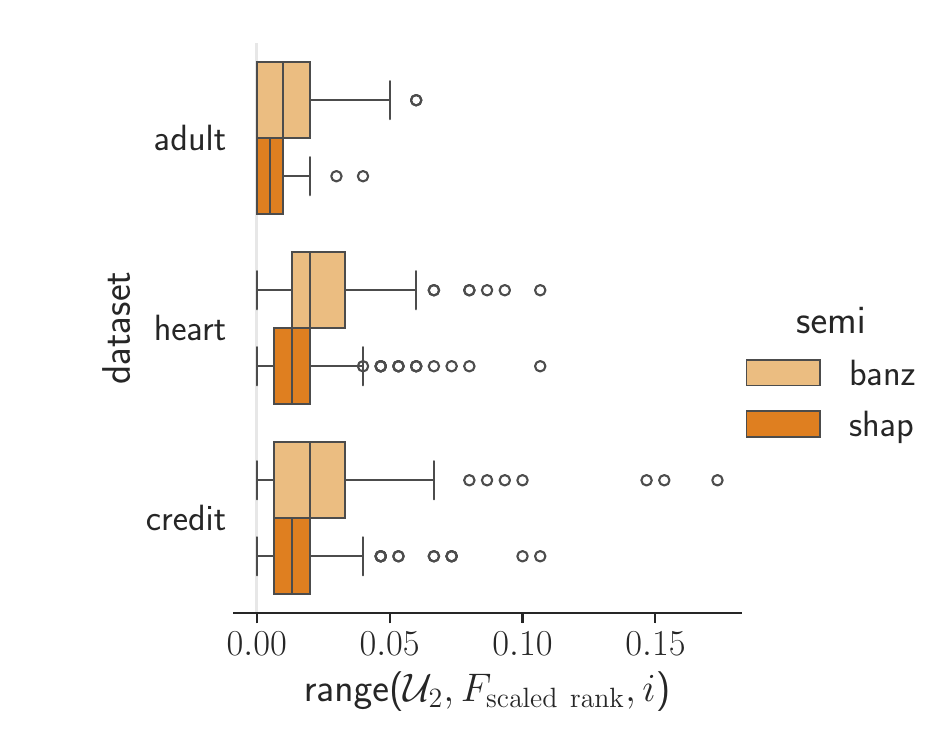}
        \caption{Cost-Ratio Parameter}
    \end{subfigure}
    \caption{Comparison of semivalue notions (Shapley vs. Banzhaf) in terms of the distribution of individual observations' scaled rank sensitivity to each explored candidate alternative set (Sec.~\ref{sec:experiments}).}%
\end{figure}\label{fig:rank_compare}

Figure~\ref{fig:rank_compare} depicts the variability of the scaled rank favorability outcome for each single-choice candidate sets under both Shapley and Banzhaf valuations. The variability  \(\mathrm{range}(\mathcal{U}, F_{\mathrm{scaled~rank}}, i)\) indicates the difference between an individual observation's worst- and best-case rank under selection of a utility function from \(\U\). If an individual can be ranked in the 75th percentile under one utility function in \(\U\) (but no better) and in the 25th percentile under another utility function in \(\U\) (but no worse), then this individual has a scaled rank with \(\mathrm{range}=0.5\) under this candidate set. 

Experiments revealed that while all datasets contain some points that are rank-stable or very near rank-stable under each of the three studied utility choices (when perturbed alone), the median observation can see significant rank shifts due to utility choice. This rank instability is particularly pronounced for Shapley valuation rankings under perturbation of small-coalitions handling. The Banzhaf value is comparably very rank stable under this candidate set due to the comparably lower weight it places on small coalition sizes. However, for monotonic $V$ transformation and cost-ratio parameter ambiguity set, the Banzhaf value shows rank instability comparable to that of the Shapley value.

Experimental results also indicate that these rank shifts can be meaningful in the context of a bottom 10\% filter. Across all datasets and candidate sets, between 2\% and 5\% of observations have their outcome under a “drop bottom 10\%” filter sensitive to utility selection.

%% file: 05_discussion.tex
Semivalue-based valuations, like all valuation methods, are fundamentally subjective. This subjectivity is often obscured by the formal appeal of semivalues: once a utility function is specified, the resulting credit allocations appear principled and precise. This perception stems from the fact that semivalues satisfy several desirable axiomatic properties, such as symmetry, linearity, and marginal contribution, that lend them normative weight. However, these axioms only gain practical legitimacy through stakeholder agreement, particularly in how counterfactual outcomes are defined and scored. Without such consensus, the fairness attributed to semivalue-based allocations is not intrinsic, but contingent on modeling choices that are often unexamined and highly contestable.

This work identifies and formalizes key sources of arbitrariness in semivalue specification and characterizes how they influence downstream outcomes. In doing so, it highlights the often-overlooked fact that even when estimators are statistically equivalent in terms of generalization or inference, they may produce vastly different valuations under small changes to the utility function. This disconnect between inferential equivalence and valuation divergence reveals a critical but subtle design space, one that complicates the assumption that semivalue-based methods can be used as neutral, objective instruments for data valuation.
Moreover, this work introduces algorithmic strategies for navigating utility ambiguity and examines how these strategies affect valuation stability. Several promising directions remain for expanding on these initial insights:

\textbf{Robustness via gameability.} While the susceptibility of valuation schemes to manipulation is often framed as a weakness, it may also present an opportunity. If adversarial perturbations or targeted utility shifts can reveal brittle or unstable regions of the valuation landscape, they may serve as diagnostic tools for robustness. Future work could formalize how bounded adversarial utility perturbations affect valuation rankings, and whether some forms of “gameability” correlate with informative sensitivities or undesirable biases. %

%\vspace{-.2em}
\textbf{Broader ambiguity sets.} The ambiguity sets considered in this work, pertaining primarily to model score functions and learning algorithms, represent only part of the space of underspecification. Other specification choices, such as whether or how to include hyperparameter tuning in counterfactuals, how approximation algorithms are implemented, and how data is preprocessed or partitioned, can further affect resulting valuations. Extending the framework to incorporate these additional sources of modeling variation could yield a more complete picture of valuation instability. %

%\vspace{-.2em}
\textbf{Extension to feature Shapley values.} Although this work focuses on data valuation, the issues of underspecification and ambiguity are not unique to it. Feature-based Shapley values, often used for model interpretability, also depend on choices, such as how to marginalize or null out features, that are rarely standardized across applications. These choices can dramatically affect interpretability claims. Adapting the analysis presented here to the feature attribution setting could clarify how modeling assumptions shape both interpretability and accountability in practice. %

%\vspace{-.2em}
\textbf{Addressing gameability.} If valuation schemes are to be used in consequential settings, such as data licensing, audit, or compensation, questions of strategic manipulation become unavoidable. Transparency alone may be insufficient if the utility function itself remains unconstrained. This raises the need for governance mechanisms that can constrain or vet utility specification. Possible avenues include provenance-tracked utility pipelines, consensus-driven specification processes, or domain-specific regulatory standards that limit the space of acceptable modeling choices.

%% file: appendices/appendix_background.tex
\section{Further Background: Semivalues}
\label{app:further_background}
Let \(N := \{1, \ldots, n\}\) be a set of players.
In a cooperative game represented as \((N, U)\), every subset of \(N\) (termed a \emph{coalition}) is assigned a value through a \emph{utility function} \(U : 2^{N} \to \mathbb{R}\).
In this setup, it is possible to map or redistribute values assigned to coalitions to the players in that coalition in multiple ways; such a redistribution is termed a \emph{payoff distribution}. 
The possibilities for the payoff distribution (or \emph{value vector}) can be restricted by placing axioms over the cooperative game; in particular, these axioms constrain how the value vector is related to the utility function \(U\), and this can result in none or more than one value vectors.
For instance, the \emph{core} is a collection of value vectors based on three axioms, and this collection can be empty, singleton or otherwise depending on the utility function.

\paragraph{Semivalues}
Given a set of players \(N\), a utility function \(U\), and a collection of non-negative weights \(\{w_{j}\}_{j = 0}^{n - 1}\), a semivalue \(\psi(U, w) \in \mathbb{R}^{n}\) is defined as
\begin{equation}
\label{eqn:semivalue-def}
    \psi(U, w)_{j} := \sum_{S \subset 2^{N \setminus \{p\}}} w_{|S|} \cdot \left\{U(S \cup \{j\}) - U(S)\right\}
\end{equation}
where \(\sum_{S \subset 2^{N \setminus \{j\}}} w_{|S|} = 1\).
A semivalue \(\psi(U, w)\) is a value vector corresponding to the cooperative game \((N, U)\) that satisfies the following three axioms.
\begin{itemize}[leftmargin=*, itemsep=0pt]
\item {\em Linearity}: for any utility functions \(U_{1}, U_{2}\) and constants \(\alpha, \beta \in \mathbb{R}\), \(\psi(\alpha U_{1} + \beta U_{2}) = \alpha \cdot \psi(U_{1}) + \beta \cdot \psi(U_{2})\).
\item {\em Anonymity}: for any permutation \(\pi\) of \(N\), let \(U^{\pi}(\pi(S)) = U(S)\) for \(S \in 2^{N}\). Then, \(\psi(U^{\pi})_{\pi(j)} = \psi(U)_{j}\).
\item {\em Dummy}: if \(j \in N\) such that \(U(S \cup \{j\}) - U(S) = U(\{j\})\) for all \(S \in 2^{N \setminus \{j\}}\), then \(\psi(U)_{j} = U(\{j\})\).
\end{itemize}
\vspace{-1em}
\begin{table}[H]
\caption{Examples of semivalues for different choices of weights}
\label{tab:example-semivalue}
\begin{center}
\begin{tabular}{cc}
Semivalue & \(w_{j}\) \\
\hline
\\
[-1em]
Shapley \citep{shapley_value_1952} & \(( n \cdot \binom{n - 1}{j})^{-1}\) \\
Banzhaf \citep{wang_banzhaf_2023} & \(2^{1 - n}\) \\
Leave-one-out (or) marginal & \(1\{j = n - 1\}\) \\
\end{tabular}
\end{center}
\end{table}
\vspace{-1em}
The distinction between the Shapley and Banzhaf value is that the Shapley uniformly weighs strata of coalition sizes, while the Banzhaf value places a uniform weight over every subset.
The Shapley value is also a special value due to the following equivalence.
\begin{talign*}
        \psi(U, w) \text{ satisfies }
        \left\{\substack{\textsf{Add}, \textsf{Sym}, \\\textsf{NP}, \textsf{Eff}}\right\} ~ \Leftrightarrow ~ w_{j} =  \left(n \cdot \binom{n - 1}{j}\right)^{-1}
\end{talign*}
and \textsf{Add}, \textsf{Sym}, \textsf{NP}, and \textsf{Eff} are defined as
\begin{itemize}[leftmargin=*, itemsep=0pt]
\item \textsf{Add}: this is a special case of linearity with \(\alpha = \beta = 1\).
\item \textsf{Sym}: suppose \(j, k\) are such that \(U(S \cup \{j\}) = U(S \cup \{k\})\) for all \(S \in 2^{N \setminus \{j, k\}}\). Then, \(\psi(U, w)_{j} = \psi(U, w)_{k}\).
\item \textsf{NP}: if \(j \in N\) such that \(U(S \cup \{j\}) = U(S)\) for all \(S \in 2^{N \setminus \{j\}}\), then \(\psi(U, w)_{j} = 0\).
\item \textsf{Eff}: \(\sum_{j=1}^{n}\psi(U, w)_{j} = U(N) - U(\emptyset)\).
\end{itemize}

\textsf{Sym} and \textsf{NP} are implied by the Anonymity and Dummy axioms respectively, but the converse is not necessarily true.
The \textsf{Eff} axiom is specific to the Shapley value; most cooperative games operate under the assumption that \(U(\emptyset) = 0\).
 \subsection{Data valuation as a cooperative game}

In the data valuation setting, we are given a dataset \(\mathcal{D} = \{z_j\}_{j=1}^{n}\) and seek to assign a value to each \(z_j\). To frame this as a cooperative game, we define a utility function that assigns a utility to every subset \(S \subseteq \mathcal{D}\). As in prior work, this function is composed of: (1) a statistical algorithm \(\mathcal{A} : 2^{\mathcal{D}} \to \mathcal{F}\) mapping subsets to a hypothesis class \(\mathcal{F}\), and (2) a performance metric \(V : \mathcal{F} \to \mathbb{R}\) that evaluates a hypothesis \(f \in \mathcal{F}\), producing a \emph{model score}. The utility function is thus defined as \( U := V \circ \mathcal{A} \), completing the cooperative game formulation.  
Assigning values to data points requires choosing axioms that (1) are well-suited to data valuation and (2) do not result in an empty set of possibilities. We focus on data semivalues, though non-semivalue approaches have also been explored \citep{yan_core_2021}.  

\subsubsection{Semivalues for Data Valuation} 

Given \(\mathcal{D}\) and \(U\), several semivalues have been proposed, including leave-one-out, Data Shapley \citep{ghorbani_shapley_2019}, Data Banzhaf \citep{wang_banzhaf_2023}, and Beta-Shapley \citep{kwon_beta_2022}. These semivalues are defined axiomatically, making it essential to assess the relevance of these axioms for data valuation:
\vspace{-.5em}
\begin{itemize}[leftmargin=*, itemsep=0pt]
\item {\em Linearity}: if multiple utility functions \(\{U_i\}_{i=1}^{m}\) are defined as \( U_i = V_i \circ \mathcal{A}_i \), linearity allows (1) parallel computation across models and (2) seamless aggregation into a single semivalue. Here, \(V_i\) correspond to an evaluation of a hypothesis over a specific test set, and each \(\mathcal{A}_i\) could represent different learning algorithms.  
\item {\em Anonymity}: ensures that a data point's value is independent of its position in the dataset. Since statistical learning algorithms often assume exchangeable data, this property naturally holds and should be reflected in valuation.  
\item {\em Dummy}: states that if a data point \( z_j \) does not impact the utility of any subset, it should receive a value of zero. However, this relies on the assumption that utility contributions are additive—an assumption often violated in machine learning, where models exhibit complex interactions between data points.  
\end{itemize} 

More recently, Shapley-based semivalues have gained attention \citep{ghorbani_shapley_2019}. Since these satisfy linearity and anonymity, their corresponding axioms, \textsf{Add} and \textsf{Sym}, hold automatically. We focus instead on two additional axioms:  
\begin{itemize}[leftmargin=*, itemsep=0pt]
\item \textsf{NP}: A point \(z_j \in \mathcal{D}\) that does not change the utility of any subset should receive a value of zero. Such points are still preferable to those that reduce coalition utility when included.  
\item \textsf{Eff}: Its relevance in machine learning has been debated. \citet{jia_shapley_2019} argue that since Shapley values do not necessarily represent a monetary reward, enforcing this mechanism is unnecessary. Similarly, \citet{kwon_beta_2022} suggest rank-based valuation mechanisms.
\end{itemize}
\vspace{-.6em}

%% file: appendices/appendix_experimental.tex
\section{Additional Experimental Details}\label{app:exp_details}

\subsection{Datasets and Baseline Learning Algorithms}
Experimental methods were adapted from code and datasets provided by the \verb|opendataval| package~\citep{jiang_opendataval_2023}. The open-source package was augmented with several additional performance metrics and a stratified sampler adapted from~\citep{wu_variance_2023}. All datasets except `birthweight' are included in the package. The `Birthweight' (lowbwt) dataset is originally from~\cite{hosmer_logistic_2000}.

Target variable for regression datasets were standardized. No additional modifications were made to the datasets.

The learning algorithm  used for classification tasks was the default logistic regression algorithm incorporated into the OpenDataVal package via \verb|ModelFactory|, with the minimum cardinality adjusted down to 1.

The learning algorithm  used for regression tasks used the OpenDataVal sklearn regressor wrapper around the default sklearn ridge regression model via \verb|ModelFactory|, with the minimum cardinality adjusted down to 1.

Essential experiments can be scaled to run on a personal laptop with 16GB memory and take less than 1 day.

\subsection{Additional Results}
\begin{figure}[h]
    \centering
    \begin{subfigure}[t]{0.32\linewidth}
        \includegraphics[width=1.1\linewidth]{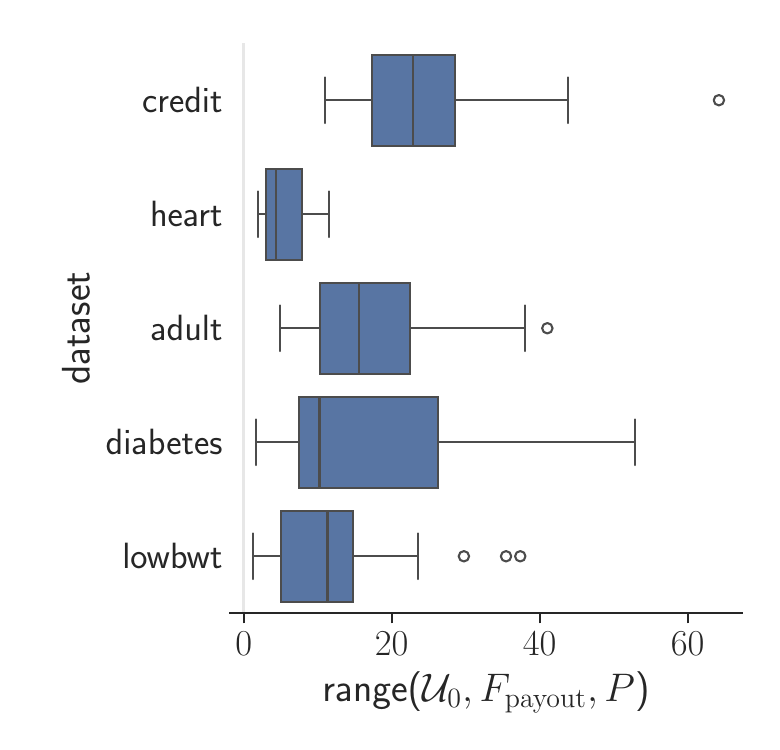}
        \caption{Under small coalition handling}
    \end{subfigure}
    \hfill
    \begin{subfigure}[t]{0.32\linewidth}
        \includegraphics[width=1.1\linewidth]{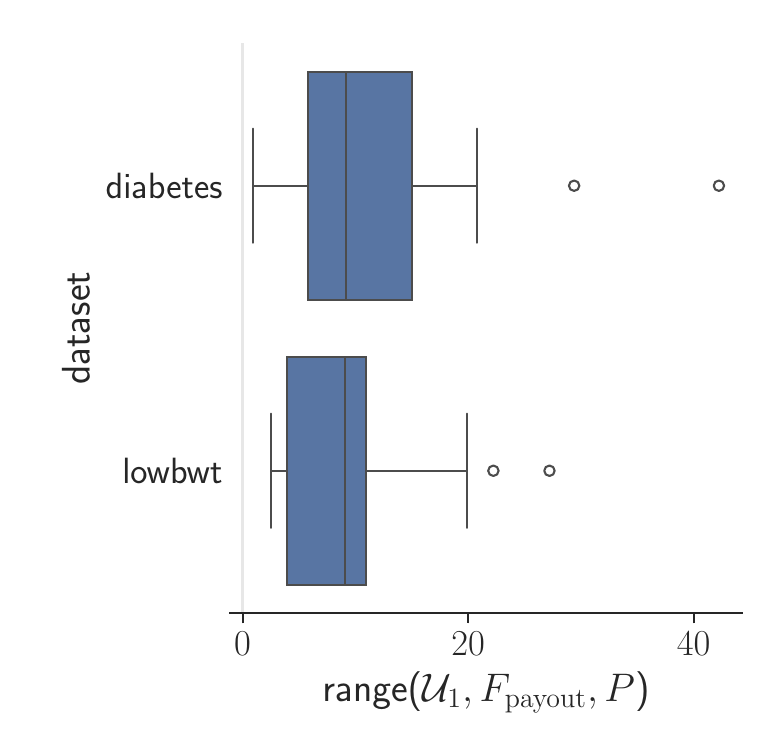}
        \caption{Monotonic \(V\) transformation}
    \end{subfigure}
    \hfill
    \begin{subfigure}[t]{0.32\linewidth}
        \includegraphics[width=1.1\linewidth]{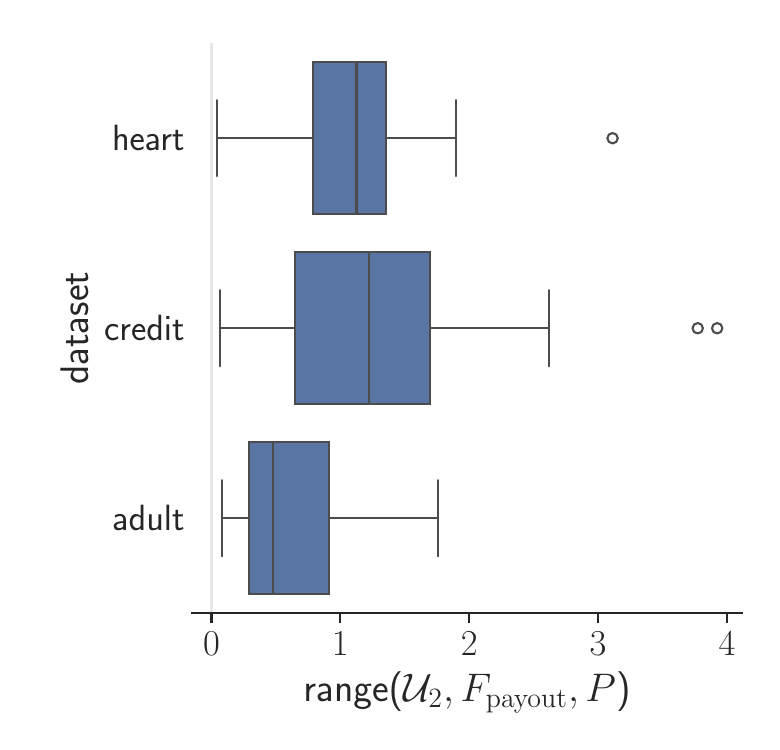}
        \caption{Cost-Ratio Parameter}
    \end{subfigure}
        \caption{Distribution of aggregate payout sensitivity to each explored candidate alternative set (\ref{sec:experiments}) for 100 randomly sampled subsets containing 10\% of observations in a dataset.}
    \label{fig:vt_gp}
\end{figure}

Figure~\ref{fig:vt_gp} provides the distribution of value shifts aggregated by group for randomly selected subsets of the data.

%% file: appendices/appendix_algorithms.tex
\newcommand{\budget}{m_{ik}}
\section{Sampling Procedures and Gameability Algorithms}\label{app:sampling}

Algorithm \ref{alg:sampler} is a stratified sampling algorithm for computing semivalues, based on \cite{maleki_bounding_2014} and \cite{ wu_variance_2023}, for evaluation of the semivalue defined by weights \(w_k\) for each individual observation in dataset \(\D\) under utility function \(U\). The algorithm takes in a budget specification \(m_{i,k}\) for \(i\in\D, k=\{0,...,N-1\}\) denoting the number of samples to be used to estimate the average marginal contribution of observation \(i\) to coalitions of size \(k\) not containing \(i\). See \citep{wu_variance_2023} for optimal budget assignment scheme for computation of Shapley values; optimal budget assignment is out of the scope of this paper.

Let \(\mathcal{S}_k^{(-i)}\) denote the stratum of coalitions of size \(k\) not including \(i\), and and let \(\mathcal{S}_k^{(+i)}\) denote the stratum of coalitions that result from adding \(i\) to these coalitions.
\[\mathcal{S}_k^{(-i)} = \{S\subset D \bigm| i\notin S,|S|=k\},\quad\quad\mathcal{S}_k^{(+i)} = \{S \cup\{i\},\forall S\in \mathcal{S}_k^{(-i)}\}.\]

Note that coalitions in \(\mathcal{S}_k^{(+i)}\) in fact have cardinality \(k+1\); this is an intentional choice so that the subscript \(k\) corresponds to the weight \(w_k\) that multiplies the utility of such coalitions in the computation of semivalues for \(i\).

In addition to computing approximate semivalues, Algorithm~\ref{alg:sampler} returns the intermediate computation products corresponding to 
 \begin{equation}\label{eq:stratum_averages}
 X_{i,k}^{-} = \mathop{\mathbb{E}}_{S\sim\mathrm{Unif}[\mathcal{S}_k^{(-i)}]}[U(S)]
 \quad\quad
  X_{i,k}^{+} = \mathop{\mathbb{E}}_{S\sim\mathrm{Unif}[\mathcal{S}_k^{(+i)}]}[U(S)]
 \end{equation}

\begin{algorithm}[ht!]
	\caption{Stratified Marginal Improvement Sampler}
	\begin{algorithmic}[1]
	\REQUIRE
	    
        Training data  $\D=\{(x_i,y_i)\}_{i=1}^N$, 

        Utility Function $U$,%Learning algorithm $\mathcal{A}$, performance metric $V$
        
        Semivalue notion \(w_k\),\\
        
        Sampling budget $\budget$ for each \(i\in\D\) and each cardinality $k\in\{0,...N-1\}$\vspace{0.1cm};\\
    \ENSURE \(\hat{\psi}_i(U,w_k), \hat{X}^{-}_{i,k}, \hat{X}^{+}_{i,k}\): estimates for semivalue defined by weights \(w_k\) for each observation \(i\in\mathcal{D}\), as well as intermediate estimates of expected stratum utilities as defined in Eq~\ref{eq:stratum_averages};\vspace{0.1cm}\\
    \STATE Initialize $\hat{X}^{+}_{i,k}, \hat{X}^{-}_{i,k}=0$\\
    \FOR{each observation $i\in\mathcal{D}$}
    \FOR{each coalition cardinality $k\in\{0,...N-1\}$}
        \FOR{$\budget$ samples}
                \STATE $S:=$ uniform sample from $\{S\subset D \bigm| i\notin S,|S|=k\}$
                 \STATE $\hat{X}^-_{i,k} = \hat{X}^-_{i,k} + \frac{1}{\budget}U(S)$;\quad\quad (compute utility contribution to  estimate of \(X^-_{i,k}\) )
                \STATE $\hat{X}^+_{i,k} = \hat{X}^+_{i,k} + \frac{1}{\budget}U(S\cup\{i\})$; \quad (compute corresponding contrib. to est. of \(X^-_{i,k}\) )
               
                \ENDFOR
        \ENDFOR
        \STATE \(\hat{\psi}_i(U,w_k) =\sum_{k=0}^{N-1}w_k (X_{i,k}^+-X_{i,k}^-)\); \quad (compute semivalue estimate)
    \ENDFOR
    \RETURN   \(\hat{\psi}_i(U,w_k), \hat{X}^{-}_{i,k}, \hat{X}^{+}_{i,k}\)
    \end{algorithmic}\label{alg:sampler}
\end{algorithm}

\clearpage
Algorithm \ref{alg:discrete_search} attempts to optimize a given favorability measure via a linear search that involves semivalue evaluation (approximate or exact) under each utility in a discrete candidate set and the computation of the favorability of that outcome. 

See Appendix~\ref{app:approx_game} for analysis of a case where this offers approximate gameability per Def.~\ref{def:approx-gameability}.

\begin{algorithm}[ht!]
	\caption{Linear search over discrete $\mathcal{U}$}
	\label{alg:sampling_value}
	\begin{algorithmic}[1]
	\REQUIRE
	    
        Training data  $\D=\{(x_i,y_i)\}_{i=1}^N$, finite set of utility candidates $\mathcal{U}$

        Semivalue computation strategy: \verb|exact| (enumeration of coalitions) or \verb|approximate| (stratified sampling, Alg~\ref{alg:sampler}, with sampling budget \(\budget\))

        Semivalue weights $w_k$,

        Favorability score $F$, preferred group $P$\\

    \ENSURE \(\hat{U}_{best},\hat{U}_{worst}\in\U\): best-guess  utility candidates that  optimize a specified favorability score \(F\) for subpopulation \(P\). 
    \STATE $\verb|max|=-\infty, \verb|min|=\infty$
    \FOR{$U\in\mathcal{U}$}
    \IF{ $\texttt{exact}$ computation of semivalues}
    \STATE  \(\psi(U, w_k)\leftarrow \) exact semivalues by enumerating all coalitions and scoring under \(U\)
    \STATE \(\verb|current| =F(\psi(U,w_k), P) \) 
    \ELSIF{ $\texttt{approximate}$ evaluation of semivalues}
    \STATE  \(\hat\psi(U, w_k)\leftarrow \) estimates \(\hat\psi(U,w_k)\) from Alg~\ref{alg:sampler} with inputs $\mathcal{D}, U, w_k, \budget$
    \STATE \(\verb|current| =F(\hat\psi(U,w_k), P) \) 
    \ENDIF
    \STATE If \(\verb|current|>\verb|max|\): update \(\hat{U}_{best}\leftarrow U\) and \(\verb|max|\leftarrow \verb|current|\);
    \STATE  If \(\verb|current|<\verb|min|\): update \(\hat{U}_{worst}\leftarrow U\) and \(\verb|min|\leftarrow \verb|current|\);
    \ENDFOR

    \RETURN \(U_{best},U_{worst}\)
    \end{algorithmic}\label{alg:discrete_search}
\end{algorithm}

\clearpage
Algorithm \ref{alg:min_k_search} seeks the most and least favorable small-coalition threshold for \(\mathcal{U}_0(k^*)\), the set of candidate utilities with an untrained model fallback rule for some \(k_{\min}\leq k^*\) (Def.~\ref{def:untrained-fallback}), under favorability score \(F\) for subpopulation \(P\). Despite the set \(\mathcal{U}_0(k^*)\) having cardinality that scales with the size of the dataset score, Algorithm~\ref{alg:min_k_search} demonstrates how using memoized results from stratified sampling can reduce complexity. 
\begin{algorithm}[ht!]
	\caption{Get approximate valuation outcomes over $\mathcal{U}_0(k^*)$}
	\begin{algorithmic}[1]
	\REQUIRE
	    
        Training data  $\D=\{(x_i,y_i)\}_{i=1}^N$, baseline utility $U_0$ %\mathcal{A}_0$, performance metric $V_0$,

        Semivalue weights $w_k$,

        Sampling budgets \(\budget\),
        
        Maximum `small cardinality' threshold $k^*$,

        Favorability score \(F\), preferenced group \(P\)
        
        %Sampling budget $b_k$ for  $k\in\{1,...N-1\}$.\\

    \ENSURE \(\hat{U}_{best}, \hat{U}_{worst}\in \mathcal{U}_0(k*)\), estimated best and worst utilities w.r.t.  favorability \(F\) for group $P$ 
    \STATE \(\verb|max|=-\infty, \verb|min|=\infty\)
    \STATE Use Alg~\ref{alg:sampler} to get  \(\hat{\psi}_i(U_0,w_k), \hat{X}^{-}_{i,k}, \hat{X}^{+}_{i,k}\) with inputs \(\mathcal{D},U_0,w_k, \budget\) 
    \FOR{$k_{min}=0,1,2,...,k^*$}
    \STATE \(U_{k_{\min}}(S) \leftarrow U_0(\emptyset)\mathbf{1}(|S|< k_{\min})+U_0(S)\mathbf{1}(|S|\geq k_{\min})\)
    \FOR{$i=1,...,N$}
        \STATE $Z_{i,k}^+ = X_{i,k}^+$ if $k\geq k_{min}-1$, otherwise $U_0(\emptyset)$
        \STATE $Z_{i,k}^- = X_{i,k}^-$ if $k\geq k_{min}$, otherwise $U_0(\emptyset)$
        \STATE $\hat{\psi}_i=\sum_{k=0}^{N-1}w_k \binom{N-1}{k}\left(\hat{Z}^+_{i,k}-\hat{Z}^-_{i,k}\right)$
    \ENDFOR
    \STATE  If \(F(\hat\psi,P)>\verb|max|\): \(\hat{U}_{best} \leftarrow U_{k_{min}}\) , \(\verb|max|\leftarrow F(\hat\psi,P)\) 
    \STATE  If \(F(\hat\psi,P)<\verb|min|\): \(\hat{U}_{worst} \leftarrow U_{k_{min}}\) , \(\verb|min|\leftarrow F(\hat\psi,P)\) 
    \ENDFOR
    \RETURN \(\hat{U}_{best}, \hat{U}_{worst}\)
    \end{algorithmic}\label{alg:min_k_search}
\end{algorithm}

\clearpage
Algorithm~\ref{alg:k_behavior_optimization} provides a way to optimize favorability \(F\) for group \(P\) when the small-coalition behavior of learning algorithm \(\alg\) is underspecified, but can be described by a discrete set of behaviors for each cardinality. 

\begin{algorithm}[ht!]
	\caption{Finite Cardinality-Dependent Behavior Gaming}
	\begin{algorithmic}[1]
	\REQUIRE
	    
        Training data  $\D=\{(x_i,y_i)\}_{i=1}^N$, learning algorithm $\mathcal{A}$ (with well specified behavior on coalitions of size \(k^*\) and above), performance metric $V$.

        Behavior options \(\{\mathcal{A}_k^{(1)},\mathcal{A}_k^{(2)},...\}\) for each \(k<k^*\).

        Semivalue weights $w_k$,

        Budget per stratum $m_{k}$
    
        Preferenced group \(P\)

    \ENSURE \(\hat{U}_{best}, \hat{U}_{worst}\), estimated best and worst small-coalition behavior selection for group \(P\). 
    
    \FOR{$k=0,1,2,...,k^*$ (will optimize behavior for each underspecified cardinality)}

    \FOR{$l=0,...,k$ (will stratify utility samples based on membership of P)}

    \STATE Compute the fraction of size-\(k\) coalitions that contain exactly \(l\) elements of \(P\):\\\quad\quad \(f_{k,l} = \binom{P}{l}\binom{\D\setminus P}{k-l}\left(\binom{N}{k}\right)^{-1}\) 
    
    \IF{cardinality \(k\) coalitions with \(l\) elements of \(P\) exist, $f_{k,l}>0$}

        \STATE Assign proportional share of sample budget the sample budget \(m_k\) for these coalitions,  \\\quad \quad\(m_{k,l}=m_{k}*f_{k,l}\)
        \STATE Sample \(\{S_j\}_{j=1}^{m_{k,l}}\) coalitions independently from the uniform distribution over coalitions with \(k\) elements containing exactly  \(l\) elements of \(P\) 
        \FOR{each allowed behavior $b$ in \(\{\mathcal{A}_k^{(1)},\mathcal{A}_k^{(2)},...\}\)}

        \STATE  Compute contribution to mean stratum utility from these samples\\ \quad \(\hat{u}_{k,l}(b)=\frac{1}{m_{k,l}}\sum_{j=1}^{m_{k,l}}(V\circ\mathcal{A}_k^{(b)})(S_j)\) 
        
    \ENDFOR

    \ENDIF

    \ENDFOR
    \STATE \(\verb|min|=\infty, \verb|max|=-\infty\)
    \FOR{each allowed behavior $b$ in \(\{\mathcal{A}_k^{(1)},\mathcal{A}_k^{(2)},...\}\)}
        \STATE Compute full stratum utility for cardinality-\(k\) coalitions:\\\quad\quad \(\hat{u}_{k}(b)=\sum_{l=0}^{k}\left(lw_k -(k-l)w_{k-1}\right) (f_{k,l})\hat{u}_{k,l}(b)\)
        \STATE If \(\hat{u}_{k}(b)>\verb|max|\): \(b_{\max} \leftarrow b\) , \(\verb|max|\leftarrow \hat{u}_b\) 
        \STATE If \(\hat{u}_{k}(b)<\verb|min|\): \(b_{\min} \leftarrow b\) , \(\verb|min|\leftarrow \hat{u}_b\) 
    \ENDFOR
    \STATE Incorporate \(\mathcal{A}_{k}^{(b_{\max})}\) as the best-case behavior for cardinality \(k\) into \(\hat{U}_{best}\)
    \STATE Incorporate \(\mathcal{A}_{k}^{(b_{\min})}\) as the worst-case behavior for cardinality \(k\) into \(\hat{U}_{worst}\)

    \ENDFOR 
    \RETURN \(\hat{U}_{best}, \hat{U}_{worst}\)
    \end{algorithmic}\label{alg:k_behavior_optimization}
\end{algorithm}
\clearpage
Algorithm \ref{alg:net_benefit_game} provides an algorithm for approximating \(U\in\U_2\), the utility candidate set arising from uncertainty in the cost-ratio parameter for a net-benefit performance function for \(p_t\in[a,b]\), that optimizes any linear combination favorability metric.

\begin{algorithm}[ht!]
	\caption{Optimization over $\mathcal{U}_{\mathrm{cost}}(a,b)$}
	\begin{algorithmic}[1]
	\REQUIRE
	    
        Training data  $\D=\{(x_i,y_i)\}_{i=1}^N$, learning task \(\alg\), test set \(\testset\),

        Maximum and minimum for net-benefit parameter \(a,b\)

        Semivalue weights $w_k$,

        Linear combination outcome function $F$

    \ENSURE \(\hat{U}_{best},\hat{U}_{worst}\in\U_2\) 
    
    \STATE Let \(U_F(S)\) compute the fraction of the test set \(\testset\) classified as false positive by model output by \(\alg\) given data \(S\). Let \(U_T(S)\) be analogously the true positive rate.
    \STATE Use Algorithm \ref{alg:sampler}  to compute \(\hat{\psi}(U_T,w_k)\) and then to compute \(\hat{\psi}(U_F,w_k)\).
    \STATE Evaluate \(F(\hat{\psi}(U_{p_t}),w_k)=\sum_{i\in \D}c_i (\hat\psi_i(U_T;w_k)+\frac{p_t}{1-p_t}\hat{\psi}_i(U_F;w_k))\) for \(p_t=a,b\)

    \RETURN \(\hat{U}_{best}=U_{p_{min}},\hat{U}_{worst}=U_{p_{max}}\) if \(F(\hat{\psi}(U_{p_{min}}),w_k)\geq F(\hat{\psi}(U_{p_{max}}),w_k)\), otherwise return \(\hat{U}_{best}=U_{p_{max}},\hat{U}_{worst}=U_{p_{min}}\)
    
    \end{algorithmic}\label{alg:net_benefit_game}
\end{algorithm}

%% file: appendices/appendix_proofs.tex
\renewcommand{\budget}{m_{ik}}
\section{Theorem Proofs}\label{app:approx_game}

\subsection{Preliminaries}
This section introduces several definitions, lemmas, and assumptions that will be used in proofs for proposition from the text.

\newcommand{\oneapprox}{(\epsilon_1,\delta_1)}
\newcommand{\F}{F(\psi(U))}
\newcommand{\hatF}{F(\hat\psi(U))}

As in Appendix~\ref{app:sampling}, we use the following notation to represent strata of subsets of \(\D\), repeated below for ease of reference.

\setcounter{definition}{7}
\begin{definition}[Coalition Strata]
Let \(\mathcal{S}_k^{(-i)}\) denote the stratum of coalitions of size \(k\) not including \(i\),
\[\mathcal{S}_k^{(-i)} = \{S\subset D \bigm| i\notin S,|S|=k\},\]
and let \(\mathcal{S}_k^{(+i)}\) denote the stratum of coalitions that result from adding \(i\) to coalitions in \(\mathcal{S}_k^{(-i)}\),
\[\mathcal{S}_k^{(+i)} = \{S \cup\{i\}:\forall S\in \mathcal{S}_k^{(-i)}\}.\]
\end{definition}

\begin{definition}[Stratum Variance of Marginal Contribution]
We denote the variance of the marginal contribution, under utility \(U\), of observations \(i\) to coalitions drawn uniformly from cardinality-\(k\) subsets of \(\D\) that exclude \(i\) as
\[
\sigma_{i,k}^2(U) =\mathop{\mathrm{var}}_{S\sim\mathrm{Unif}[\mathcal{S}_k^{(-i)}]}[U(S\cup\{i\})-U(S)].
\]

\end{definition}

Assumption 1 employs a set of utility candidates with a bounded range to provide a straightforward condition in which stratum variance is bounded and does not carry dependence on the dataset size. However, standard stratified sampling procedures still offer the runtime required by the gameability definition under the weaker condition:

\begin{assumption} For any considered \(U\), for a fixed task on a dataset drawn from some population, \(\max_{i,k}[\sigma_{ik}(U)]\) scales at worst as polynomial in dataset size. 
\end{assumption}

As discussed in the text, many performance scores are defined on a fixed range (such as accuracy, precision, recall, AUCROC). However, in the cases where no such bounds are readily available, Since practical learning algorithms under normal condition converge as data is added, we expect \(\sigma_{i,k}(U)\) to decrease with \(k\). This leaves the only concern for assumption 2 as larger datasets adding outliers that cause poor model performance when in small coalitions. For most natural data generating procedures and well-behaved learning algorithms, this is not likely to introduce unbounded variance. In cases where this cannot be verified, the option remains to clip the score function at some multiple of the error from the untrained model remains. 
See~\cite{maleki_bounding_2014}$\S$4,~\cite{ghorbani_shapley_2019} Appendix A, and~\cite{han_replication-robust_2023}$\S$V.B for further discussion of how the runtime of stratified sampling algorithms depend on the range and variance of marginal contributions.

\begin{lemma}[Behavior of Alg~\ref{alg:sampler}]\label{lem:sampler}
Given budget specification of \(\budget\) (ie, number of sampled the marginal contribution of observation \(i\) to coalitions of cardinality \(k\)), Alg.~\ref{alg:sampler} provides semivalue estimates that are unbiased,
\[\mathbb{E}[\hat\psi_{i}(U,w_k)]=\psi_i(U),\quad \forall i\in\D\]

with variance,
\[\mathrm{var}[\hat\psi_{i}(U,w_k)]= \sum_{k=0}^{N-1}\binom{N-1}{k}^2\frac{w_k^2}{\budget}\sigma^2_{i,k}(U).\]
Further, semivalues are sampled independently for each observation, 
\[\hat\psi_i\indep\hat\psi_j, \quad i\neq j\]
\end{lemma}

\textbf{Proof}

 Algorithm~\ref{alg:sampler} draws coalition samples from from each stratum \(\mathcal{S}_k^{(-i)}\) uniformly to compute the average marginal contribution of \(i\) to coalitions of size \(k\) that exclude it. All coalition samples are drawn independently.
 
 This algorithm provides unbiased estimates for each semivalue:
 
 \[\begin{aligned}\mathbb{E}[\hat\psi_{i}(U,w_k)]&=  \sum_{k=0}^{N-1}\binom{N-1}{k}\sum_{j=1}^{B}\frac{w_k}{B} \mathop{\mathbb{E}}_{S_j\sim \mathrm{Unif}[\mathcal{S}_k^{(-i)}]}\left[U(S_j\cup\{i\})-U(S_j)]\right]\\&=\sum_{k=0}^{N-1}w_k\sum_{S\in \mathcal{S}_k^{(-i)}}[{U(S\cup\{i\})-U(S)}]\\&=\psi_i(U,w_k).\end{aligned}\]

 By similar computation, again recalling that coalitions are drawn independently, 
  \[\begin{aligned}\mathrm{var}[\hat\psi_{i}(U,w_k)]
    &= \sum_{k=0}^{N-1} \mathop{\mathrm{var}}_{S_j\sim \mathrm{Unif}[\mathcal{S}_k^{(-i)}]}\left[\sum_{j=1}^{B}\binom{N-1}{k}\frac{w_k}{B}[U(S_j\cup\{i\})-U(S_j)]\right]\\
    &=\sum_{k=0}^{N-1}\binom{N-1}{k}^2\frac{w_k^2}{B} \mathop{\mathrm{var}}_{S_j\sim \mathrm{Unif}[\mathcal{S}_k^{(-i)}]}[{U(S\cup\{i\})-U(S)}]\\
    &= \sum_{k=0}^{N-1}\binom{N-1}{k}^2\frac{w_k^2}{B}\sigma_{i,k}^2(U).\end{aligned}\]

 Finally,  \(\hat{\psi}_i(U,w_k) \indep \hat{\psi}_j(U,w_k)  \) for \(i\neq j\) also arises as a consequence of the independent draws of coalitions.

\subsection{Proposition 1: Gameability of Fixed Cardinality Candidate Set}

\setcounter{proposition}{0}
\begin{proposition}[$(\epsilon, \delta)$-Gameability of Fixed Cardinality \(\mathcal{U}\)]
Let \(\mathcal{U}\) be a finite candidate set of utility functions of size \(O(1)\), i.e., independent of the dataset size \( |\mathcal{D}| = N\).
%Let \(\psi\) be any semivalue notion. 
\(\mathcal{U}\) is $(\epsilon,\delta)$-gameable under favorability \(\aggF\) for any \(P\) using Algorithm~\ref{alg:discrete_search}.
\end{proposition}

\textbf{Proof}

\newcommand{\otherterms}{K|\mathcal{U}|^2}

To simplify notation, let \(\psi(U)=\psi(U,w_k)\) and let \(F\) denote the aggregate value favorability  score for population arbitrary \(P\) based on the \(w_k\)-semivalue:
\[F(\psi(U))=F_{\mathrm{agg}}(\psi(U,w_k),P) = \sum_{i\in P}\psi_i(U,w_k).\]

Running Algorithm \ref{alg:sampler} for each \(U\in\U\) with constant budget specification, $\budget=B$, provides semivalue estimates \(\hat\psi_i(U)\) for each \(i\in\D\). Per Lemma~\ref{lem:sampler}, all estimate are unbiased, independent and with known variance. As \(F\) is the sum over semivalue estimates for each \(i\in P\), by linearity of expectation, the using the semivalue estimates to estimate favorability \(F\) provides an unbiased estimate of \(F(\psi(U))\),
\[\mathbb{E}[F(\hat{\psi}(U))] = \mathbb{E}[\sum_{i\in P}\hat\psi(U)]=\sum_{i\in P}\mathbb{E}[\hat\psi(U)]=\sum_{i\in P}\psi(U)=F(\psi(U)).\]

The variance of \(F(\hat\psi(U))\) can be computed by adding the variance of the semivalue estimate for each \(i\in P\),

\[\mathrm{var}[\hatF]= \sum_{i\in P}\sum_{k=0}^{N-1}\binom{N-1}{k}^2\frac{w_k^2}{\budget}\sigma_{i,k}(U)\]

Let \(U^*=\arg\max_{U\in\U}[F(U)]\) be the true optimal utility in \(\U\) under metric \(F\).

We consider the probability that \ref{alg:discrete_search} returns a suboptimal utility function whose favorability under \(F\) falls below the optimal favorability by more than \(\epsilon\). For this to happen requires, we must have that for some \(U'\in\mathcal{U}\),
\begin{enumerate}
    \item  \((F(\hat\psi;U')-F(\hat\psi;U^*))>0\), else it wouldn't be returned as the ideal utility by Algorithm \ref{alg:discrete_search}), and
    \item \((F(\psi;U')-F(\psi;U^*))<\epsilon\), else it has favorability within \(\epsilon\) of optimal and is appropriate to return
\end{enumerate}

As such, the probability that a specific \(\epsilon\)-suboptimal \(U'\) being returned is upper bounded by:

\[\mathrm{Pr}[F(\hat\psi,U')-F(\hat\psi,U^*)>0\bigm|\mathbb{E}[F(\hat\psi,U)-F(\hat\psi,U^*)]<-\epsilon]=\mathrm{Pr}[\mathsf{X}>\epsilon]\]

For random variable \(\mathsf{X}\) with mean 0 and variance \[\mathrm{var}(\mathsf{X})=\sum_{i\in P}\sum_{k=0}^{N-1}\binom{N-1}{k}^2\frac{w_k^2}{B}\left(\sigma^2_{i,k}(U')+\sigma^2_{i,k}(U^*)\right)=\sigma^2_{U'},\]

By the Chebyshev bound
\[\mathrm{Pr}[\text{U' returned by Alg 2}|\text{U' is } \epsilon\text{-suboptimal}]\leq \frac{\sigma^2_{U'}}{B\epsilon^2}\]

And with the union bound:
\[\mathrm{Pr}[\text{any }\epsilon\text{-suboptimal U' returned by Alg 2}]\leq \frac{|\U|\sigma^2_{U'}}{B\epsilon_0^2}\]

Adapting~\cite {maleki_bounding_2014}, this establishes that a sampling budget of \(B=O(|\mathcal{U}|^2 \log(1/\delta)\epsilon^2)\) offers an \((\epsilon,\delta)\)-approximation. With Assumption 1 and the \(|\U|\) that does not scale with dataset size, this provides the \(O(\mathrm{poly}(N)\log(1/\delta)/\epsilon^2)\) required to satisfy Definition~\ref{def:approx-gameability}.

\subsection{Proposition 2: Gameability of Small Cardinality Behavior}
\begin{proposition}[$(\epsilon, \delta)$-Gameability of Small-Cardinality Behavior]
Let \(\mathcal{U}\subseteq \mathcal{U}_{\mathrm{small}}\) be the set of utility functions within the small-cardinality ambiguity set such that the behavior of the learning algorithm can be specified by a fixed number of possible behaviors. This candidate set is \(\mathcal{U}_{0}(k^*) \) is $(\epsilon,\delta)$-gameable under favorability  \(\aggF\) for any \(P\). %
\end{proposition}

\textbf{Proof}
Let  \(\mathcal{A}\) denote a baseline learning algorithm with behavior well defined for coalitions with cardinality \(k^*\) or greater.  Consider the finite set  \(\{\mathcal{A}_k^{(2)},\mathcal{A}_k^{(1)},...,\mathcal{A}_k^{(m)} \}\) of plausible specifications for the behavior of the learning algorithm on coalitions of size exactly \(k\). \(\mathcal{A}_k^{(\cdot)}\) is defined to map subsets of \(\D\) to models that can be scored by \(V\). Presume that such behavior sets can be defined for all \(k<k^*\) and that each has cardinality that does not scale with problem size.

We define a `partial utility' from selecting behavior \(b\) for handling coalitions of size \(k\) as
\[U_k^{(b)}(S)=\begin{cases}(V\circ\mathcal{A}_k^{(b)})(S) & |S|=k\\0&\text{otherwise}\end{cases}\]

The utility of a learning algorithm that chooses behaviors \(\mathbf{b}=b_0, b_1,...b_{k^*-1}\) for handling sets of cardinality $k=1,2,\dots,k^*-1$  (for which the baseline algorithm is un- or under-defined) is thus:
\[U^\mathbf{b}(S) = \begin{cases}(V\circ \mathcal{A})(S) & |S|\geq k^* \\(V\circ \mathcal{A}_k^{(b_k)})(S) & |S|=k<k^*\end{cases}\]

\(\mathcal{U}=\{U^{\mathbf{b}}|\text{any behavior label~}\mathbf{b}\}\) is then the set of utility functions that are a subset of \(\U_{\mathrm{small}}(k^*)\) that have an allowable behavior for cardinality \(k\) coalitions. 

We now seek to optimize the aggregate value favorability for a preferenced subgroup \(P\).

The aggregate value of an arbitrary population \(P\) for semivalue with weights \(w_k\) can be computed as 
\[\aggF=\sum_{S\subseteq\mathcal{D}} \alpha (|S|,|S\cap P|)U(S)\]
where \[\alpha(k,l) = (l)w_{k-1}+(k-l)w_{k}\]
%\[\alpha(k=|S|,l=|S\cap P|) = |S\cap P|w_{k-1}+(|S|-|S\cap P|)w_{k}\]
aggregates the coefficient on the \(U(S)\) term across all individuals in P. 

The aggregate value favorability of \(U^{\mathbf{b}}\) for population \(P\) can thus be decomposed and optimized with:
\[\begin{aligned}\arg\max_{\mathbf{b}}\aggF &=\arg\max_{\mathbf{b}}\sum_{S\subseteq\mathcal{D}} \alpha (S)U^{\mathbf{b}}(S)\\&=\arg\max_{\mathbf{b}}\sum_{k=1}^{k^*-1}\sum_{l=1}^{\min(k,|P|)}\alpha (k,l)\left(\sum_{\substack{S\subseteq\mathcal{D}\\|S|=k, ~|S\cap P|=l}} U^{b_k}(S)\right)\end{aligned}\]
When behaviors can be selected independently, the above can be optimized by optimizing the behavior for each cardinality independently:
\[b_k\in\arg\max_{b_k}\sum_{l=1}^{\min(k,|P|)}\alpha (k,l)\left(\sum_{\substack{S\subseteq\mathcal{D}\\|S|=k,~|S\cap P|=l}} U^{b_k}(S)\right)\]
The optimization of behavior \(b_k\) only requires sampling coalitions of size k. This keeps the sampling complexity polynomial in dataset size, as required by Def.~\ref{def:approx-gameability}, despite the potential for exponential complexity of the set of candidate utilities arising from all possible combinations of small cardinality behaviors.

\subsection{Proposition 3: Gameability of Cost-Ratio}
\begin{proposition}[$(\epsilon,\delta)$-Gameability of $\U_{cost}$]
Let \(\mathcal{U}_{cost}(a,b)\) be a utility ambiguity set arising from Definition~\ref{def:cost-ambiguity}. \(\mathcal{U}_{cost}\) is $(\epsilon,\delta)$-gameable under favorability \(F_{\mathrm{agg}}(\psi(U);P)\) for any \(P\) with respect to  \(\psi\) using Algorithm~\ref{alg:net_benefit_game} .
\end{proposition}

\textbf{Proof} 
Consider the computation of the semivalue (with weights \(w_k\)) of observations in \(\D\) based on the performance of models learned by \(\alg\) and assessed by the net-benefit performance metric with unknown parameter \(p_t\). Let \(U_{p_t}\in \U_{\mathrm{cost}}(a, b)\) denote the utility function arising from the net benefit performance metric with parameter \(p_t\). 

From the linearity axiom of semivalues, we note that \[\psi(U_{p_T})=\psi(U_{T})+\left(\frac{1}{1-p_t}\right)\psi(U_F)\]
where \(U_T\) and \(U_F\) denote the utility function arising from learning algorithm \(\alg\) and performance metrics measuring the model's true positive and false positive rates. Noting the linearity of the aggregate reward favorability metric for arbitrary subgroup \(P\) 

\[\aggF = \sum_{i\in P}\psi_i(U_{p_t},w_k)=\sum_{i\in P}\psi_i(U_{T},w_k) + \left(\frac{p_t}{1-p_t}\right)\sum_{i\in P}\psi_i(U_{T},w_k)\]

It can be easily seen that \(F_{\mathrm{agg}}\) is linear in \((\frac{p_t}{1-p_t})\), which is a monotone function of parameter \(p_t\). Thus, extrema of \(\aggF\) over \(U\in \U_{\mathrm{cost}}(a, b)\) are achieved at the boundaries, ie \[(\arg\max_{U\in \U_{\mathrm{cost}}(a, b)}(\aggF))\cap\{U_a, U_b\}\neq\emptyset\]
 \[(\arg\min_{U\in \U_{\mathrm{cost}}(a, b)}(\aggF))\cap\{U_a, U_b\}\neq\emptyset\]

Algorithm \ref{alg:net_benefit_game} uses Algorithm \ref{alg:discrete_search} over discrete candidate set \(\{U_{a},U_b\}\). Per \ref{prop:fixed-gameability}, this offers a \(\delta,\epsilon\) solution to selecting the optimal utility in \(\U_{\mathrm{cost}}(a, b)\)

%% file: SemivalueArbitraryGameable.bbl
\begin{thebibliography}{39}
\providecommand{\natexlab}[1]{#1}
\providecommand{\url}[1]{\texttt{#1}}
\expandafter\ifx\csname urlstyle\endcsname\relax
  \providecommand{\doi}[1]{doi: #1}\else
  \providecommand{\doi}{doi: \begingroup \urlstyle{rm}\Url}\fi

\bibitem[Agarwal et~al.(2019)Agarwal, Dahleh, and Sarkar]{agarwal_marketplace_2019}
A.~Agarwal, M.~Dahleh, and T.~Sarkar.
\newblock A {Marketplace} for {Data}: {An} {Algorithmic} {Solution}, May 2019.
\newblock URL \url{http://arxiv.org/abs/1805.08125}.

\bibitem[Barocas et~al.(2023)Barocas, Hardt, and Narayanan]{barocas_fairness_ml}
S.~Barocas, M.~Hardt, and A.~Narayanan.
\newblock \emph{Fairness and Machine Learning: Limitations and Opportunities}.
\newblock MIT Press, 2023.

\bibitem[Black et~al.(2022)Black, Raghavan, and Barocas]{black_multiplicity_2022}
E.~Black, M.~Raghavan, and S.~Barocas.
\newblock Model {Multiplicity}: {Opportunities}, {Concerns}, and {Solutions}.
\newblock In \emph{2022 {ACM} {Conference} on {Fairness}, {Accountability}, and {Transparency}}, pages 850--863, Seoul Republic of Korea, June 2022. ACM.
\newblock ISBN 978-1-4503-9352-2.
\newblock \doi{10.1145/3531146.3533149}.
\newblock URL \url{https://dl.acm.org/doi/10.1145/3531146.3533149}.

\bibitem[Bornschein et~al.(2020)Bornschein, Visin, and Osindero]{bornschein_small_2020}
J.~Bornschein, F.~Visin, and S.~Osindero.
\newblock Small data, big decisions: Model selection in the small-data regime.
\newblock \emph{CoRR}, abs/2009.12583, 2020.
\newblock URL \url{https://arxiv.org/abs/2009.12583}.

\bibitem[Bower et~al.(2017)Bower, Kitchen, Niss, Strauss, Vargas, and Venkatasubramanian]{bower_fair_2017}
A.~Bower, S.~N. Kitchen, L.~Niss, M.~J. Strauss, A.~Vargas, and S.~Venkatasubramanian.
\newblock Fair {Pipelines}, July 2017.
\newblock URL \url{http://arxiv.org/abs/1707.00391}.
\newblock arXiv:1707.00391 [cs].

\bibitem[Cheong et~al.(2023)Cheong, Kim, and Vaquero]{cheong_investment_2023}
H.~Cheong, B.~Kim, and I.~U. Vaquero.
\newblock A {Data} {Valuation} {Model} to {Estimate} the {Investment} {Value} of {Platform} {Companies}: {Based} on {Discounted} {Cash} {Flow}.
\newblock \emph{Journal of Risk and Financial Management}, 16\penalty0 (6):\penalty0 293, June 2023.
\newblock ISSN 1911-8074.
\newblock \doi{10.3390/jrfm16060293}.
\newblock URL \url{https://www.mdpi.com/1911-8074/16/6/293}.
\newblock Number: 6 Publisher: Multidisciplinary Digital Publishing Institute.

\bibitem[Creel and Hellman(2022)]{creel_leviathan_2022}
K.~Creel and D.~Hellman.
\newblock The {Algorithmic} {Leviathan}: {Arbitrariness}, {Fairness}, and {Opportunity} in {Algorithmic} {Decision}-{Making} {Systems}.
\newblock \emph{Canadian Journal of Philosophy}, 52\penalty0 (1):\penalty0 26--43, Jan. 2022.
\newblock ISSN 0045-5091, 1911-0820.
\newblock \doi{10.1017/can.2022.3}.
\newblock URL \url{https://www.cambridge.org/core/product/identifier/S0045509122000030/type/journal_article}.

\bibitem[Diehl and Wilson(2025)]{diehl_arbitrary_2025}
H.~Diehl and A.~C. Wilson.
\newblock The surprising amount of arbitrariness in shapley-value data valuation.
\newblock \emph{Navigating and Addressing Data Problems for Foundation Models (DATA-FM @ ICLR)}, 2025.

\bibitem[Dwork and Ilvento(2019)]{dwork_composition_2019}
C.~Dwork and C.~Ilvento.
\newblock Fairness {Under} {Composition}.
\newblock \emph{LIPIcs, Volume 124, ITCS 2019}, 124:\penalty0 33:1--33:20, 2019.
\newblock ISSN 1868-8969.
\newblock \doi{10.4230/LIPIcs.ITCS.2019.33}.
\newblock URL \url{http://arxiv.org/abs/1806.06122}.
\newblock arXiv:1806.06122 [cs].

\bibitem[Dwork et~al.(2020)Dwork, Ilvento, Rothblum, and Sur]{dwork_abstracting_2020}
C.~Dwork, C.~Ilvento, G.~N. Rothblum, and P.~Sur.
\newblock Abstracting {Fairness}: {Oracles}, {Metrics}, and {Interpretability}, Apr. 2020.
\newblock URL \url{http://arxiv.org/abs/2004.01840}.
\newblock arXiv:2004.01840 [cs].

\bibitem[Enshaei et~al.(2021)Enshaei, Rafiee, Mohammadi, and Naderkhani]{enshaei_data_2021}
N.~Enshaei, M.~J. Rafiee, A.~Mohammadi, and F.~Naderkhani.
\newblock Data {Shapley} {Value} for {Handling} {Noisy} {Labels}: {An} application in {Screening} {COVID}-19 {Pneumonia} from {Chest} {CT} {Scans}, Oct. 2021.
\newblock URL \url{http://arxiv.org/abs/2110.08726}.
\newblock arXiv:2110.08726 [cs, eess].

\bibitem[Fabris et~al.(2022)Fabris, Messina, Silvello, and Susto]{fabris_algorithmic_2022}
A.~Fabris, S.~Messina, G.~Silvello, and G.~A. Susto.
\newblock Algorithmic {Fairness} {Datasets}: the {Story} so {Far}.
\newblock \emph{Data Mining and Knowledge Discovery}, 36\penalty0 (6):\penalty0 2074--2152, Nov. 2022.
\newblock ISSN 1384-5810, 1573-756X.
\newblock \doi{10.1007/s10618-022-00854-z}.
\newblock URL \url{http://arxiv.org/abs/2202.01711}.
\newblock arXiv:2202.01711 [cs].

\bibitem[Ganesh et~al.(2025)Ganesh, Taik, and Farnadi]{ganesh_arbitrariness_2025}
P.~Ganesh, A.~Taik, and G.~Farnadi.
\newblock The {Curious} {Case} of {Arbitrariness} in {Machine} {Learning}, Jan. 2025.
\newblock URL \url{http://arxiv.org/abs/2501.14959}.
\newblock arXiv:2501.14959 [cs] version: 1.

\bibitem[Ghorbani and Zou(2019)]{ghorbani_shapley_2019}
A.~Ghorbani and J.~Zou.
\newblock Data {Shapley}: {Equitable} {Valuation} of {Data} for {Machine} {Learning}, June 2019.

\bibitem[Ghorbani et~al.(2020)Ghorbani, Kim, and Zou]{ghorbani_distributional_2020}
A.~Ghorbani, M.~P. Kim, and J.~Zou.
\newblock A {Distributional} {Framework} for {Data} {Valuation}, Feb. 2020.
\newblock URL \url{https://arxiv.org/abs/2002.12334v1}.

\bibitem[Han et~al.(2023)Han, Wooldridge, Rogers, Ohrimenko, and Tschiatschek]{han_replication-robust_2023}
D.~Han, M.~Wooldridge, A.~Rogers, O.~Ohrimenko, and S.~Tschiatschek.
\newblock Replication-{Robust} {Payoff}-{Allocation} for {Machine} {Learning} {Data} {Markets}.
\newblock \emph{IEEE Transactions on Artificial Intelligence}, 4\penalty0 (5):\penalty0 1114--1128, Oct. 2023.
\newblock ISSN 2691-4581.
\newblock \doi{10.1109/TAI.2022.3195686}.
\newblock URL \url{http://arxiv.org/abs/2006.14583}.
\newblock arXiv:2006.14583 [cs].

\bibitem[Hosmer and Lemeshow(2000)]{hosmer_logistic_2000}
D.~Hosmer and S.~Lemeshow.
\newblock \emph{Applied Logistic Regression}.
\newblock John Wiley \& Sons, Ltd, 2000.
\newblock ISBN 9780471722144.

\bibitem[Jia et~al.(2019{\natexlab{a}})Jia, Dao, Wang, Hubis, G{\"{u}}rel, Li, Zhang, Spanos, and Song]{jia_knn_2019}
R.~Jia, D.~Dao, B.~Wang, F.~A. Hubis, N.~M. G{\"{u}}rel, B.~Li, C.~Zhang, C.~J. Spanos, and D.~Song.
\newblock Efficient task-specific data valuation for nearest neighbor algorithms.
\newblock \emph{CoRR}, abs/1908.08619, 2019{\natexlab{a}}.
\newblock URL \url{http://arxiv.org/abs/1908.08619}.

\bibitem[Jia et~al.(2019{\natexlab{b}})Jia, Dao, Wang, Hubis, Hynes, Gurel, Li, Zhang, Song, and Spanos]{jia_shapley_2019}
R.~Jia, D.~Dao, B.~Wang, F.~A. Hubis, N.~Hynes, N.~M. Gurel, B.~Li, C.~Zhang, D.~Song, and C.~Spanos.
\newblock Towards {Efficient} {Data} {Valuation} {Based} on the {Shapley} {Value}, Feb. 2019{\natexlab{b}}.
\newblock URL \url{https://arxiv.org/abs/1902.10275v4}.

\bibitem[Jiang et~al.(2023)Jiang, Liang, Zou, and Kwon]{jiang_opendataval_2023}
K.~F. Jiang, W.~Liang, J.~Zou, and Y.~Kwon.
\newblock {OpenDataVal}: a {Unified} {Benchmark} for {Data} {Valuation}, June 2023.
\newblock URL \url{http://arxiv.org/abs/2306.10577}.
\newblock arXiv:2306.10577 [cs, stat].

\bibitem[Jurcys et~al.(2020)Jurcys, Donewald, Fenwick, Lampinen, and Smaliukas]{jurcys_ownership_2020}
P.~Jurcys, C.~Donewald, M.~Fenwick, M.~Lampinen, and A.~Smaliukas.
\newblock Ownership of {User}-{Held} {Data}: {Why} {Property} {Law} {Is} the {Right} {Approach}.
\newblock \emph{SSRN Electronic Journal}, 2020.
\newblock ISSN 1556-5068.
\newblock \doi{10.2139/ssrn.3711017}.
\newblock URL \url{https://www.ssrn.com/abstract=3711017}.

\bibitem[Kleinberg et~al.(2001)Kleinberg, Papadimitriou, and Raghavan]{kleinberg_value_2001}
J.~Kleinberg, C.~H. Papadimitriou, and P.~Raghavan.
\newblock On the value of private information.
\newblock In \emph{Proceedings of the 8th conference on {Theoretical} aspects of rationality and knowledge}, {TARK} '01, pages 249--257, San Francisco, CA, USA, July 2001. Morgan Kaufmann Publishers Inc.
\newblock ISBN 978-1-55860-791-0.

\bibitem[Kumar et~al.(2022)Kumar, Lakshminarayanan, Chang, Guretno, Mien, Kalpathy-Cramer, Krishnaswamy, and Singh]{kumar_federated_2022}
S.~Kumar, A.~Lakshminarayanan, K.~Chang, F.~Guretno, I.~H. Mien, J.~Kalpathy-Cramer, P.~Krishnaswamy, and P.~Singh.
\newblock Towards {More} {Efficient} {Data} {Valuation} in {Healthcare} {Federated} {Learning} using {Ensembling}, Sept. 2022.
\newblock URL \url{http://arxiv.org/abs/2209.05424}.
\newblock arXiv:2209.05424 [cs].

\bibitem[Kwon and Zou(2022)]{kwon_beta_2022}
Y.~Kwon and J.~Zou.
\newblock Beta {Shapley}: a {Unified} and {Noise}-reduced {Data} {Valuation} {Framework} for {Machine} {Learning}, Jan. 2022.
\newblock URL \url{http://arxiv.org/abs/2110.14049}.
\newblock arXiv:2110.14049 [cs, stat].

\bibitem[Maleki et~al.(2014)Maleki, Tran-Thanh, Hines, Rahwan, and Rogers]{maleki_bounding_2014}
S.~Maleki, L.~Tran-Thanh, G.~Hines, T.~Rahwan, and A.~Rogers.
\newblock Bounding the {Estimation} {Error} of {Sampling}-based {Shapley} {Value} {Approximation}, Feb. 2014.
\newblock URL \url{http://arxiv.org/abs/1306.4265}.
\newblock arXiv:1306.4265 [cs].

\bibitem[Moody and Walsh(1999)]{moody_walsh}
D.~Moody and P.~Walsh.
\newblock Measuring {The} {Value} {Of} {Information}: {An} {Asset} {Valuation} {Approach}.
\newblock In \emph{European Conference on Information Systems}, 1999.
\newblock URL \url{https://api.semanticscholar.org/CorpusID:9136893}.

\bibitem[Namba et~al.(2024)Namba, Horiguchi, Hamamoto, and Egi]{namba_thresholding_2024}
H.~Namba, S.~Horiguchi, M.~Hamamoto, and M.~Egi.
\newblock Thresholding data shapley for data cleansing using multi-armed bandits, 2024.
\newblock URL \url{https://arxiv.org/abs/2402.08209}.

\bibitem[Schoch et~al.(2023)Schoch, Mishra, and Ji]{schoch_data_2023}
S.~Schoch, R.~Mishra, and Y.~Ji.
\newblock Data {Selection} for {Fine}-tuning {Large} {Language} {Models} {Using} {Transferred} {Shapley} {Values}, June 2023.
\newblock URL \url{http://arxiv.org/abs/2306.10165}.
\newblock arXiv:2306.10165 [cs].

\bibitem[Shapley(1952)]{shapley_value_1952}
L.~S. Shapley.
\newblock A {Value} for {N}-{Person} {Games}.
\newblock Technical report, RAND Corporation, Mar. 1952.
\newblock URL \url{https://www.rand.org/pubs/papers/P295.html}.

\bibitem[Sim et~al.(2022)Sim, Xu, and Low]{sim_data_2022}
R.~H.~L. Sim, X.~Xu, and B.~K.~H. Low.
\newblock Data {Valuation} in {Machine} {Learning}: "{Ingredients}", {Strategies}, and {Open} {Challenges}.
\newblock In \emph{Proceedings of the {Thirty}-{First} {International} {Joint} {Conference} on {Artificial} {Intelligence}}, Vienna, Austria, July 2022. International Joint Conferences on Artificial Intelligence Organization.
\newblock ISBN 978-1-956792-00-3.
\newblock \doi{10.24963/ijcai.2022/782}.
\newblock URL \url{https://www.ijcai.org/proceedings/2022/782}.

\bibitem[Tang et~al.(2021)Tang, Ghorbani, Yamashita, Rehman, Dunnmon, Zou, and Rubin]{tang_medical_2021}
S.~Tang, A.~Ghorbani, R.~Yamashita, S.~Rehman, J.~A. Dunnmon, J.~Zou, and D.~L. Rubin.
\newblock Data valuation for medical imaging using {Shapley} value and application to a large-scale chest {X}-ray dataset.
\newblock \emph{Scientific Reports}, 11\penalty0 (1), Apr. 2021.
\newblock ISSN 2045-2322.
\newblock \doi{10.1038/s41598-021-87762-2}.
\newblock URL \url{https://www.nature.com/articles/s41598-021-87762-2}.

\bibitem[Tian et~al.(2023)Tian, Liu, Li, Cao, Jia, Kong, Liu, and Ren]{tian_private_2023}
Z.~Tian, J.~Liu, J.~Li, X.~Cao, R.~Jia, J.~Kong, M.~Liu, and K.~Ren.
\newblock Private {Data} {Valuation} and {Fair} {Payment} in {Data} {Marketplaces}, Feb. 2023.
\newblock URL \url{http://arxiv.org/abs/2210.08723}.

\bibitem[Wang and Jia(2023)]{wang_banzhaf_2023}
J.~T. Wang and R.~Jia.
\newblock Data {Banzhaf}: {A} {Robust} {Data} {Valuation} {Framework} for {Machine} {Learning}, Mar. 2023.
\newblock URL \url{http://arxiv.org/abs/2205.15466}.

\bibitem[Wang et~al.(2024)Wang, Deng, Chiba-Okabe, Barak, and Su]{wang_economic_2024}
J.~T. Wang, Z.~Deng, H.~Chiba-Okabe, B.~Barak, and W.~J. Su.
\newblock An {Economic} {Solution} to {Copyright} {Challenges} of {Generative} {AI}, Sept. 2024.
\newblock URL \url{http://arxiv.org/abs/2404.13964}.
\newblock arXiv:2404.13964 [cs].

\bibitem[Wang et~al.(2020)Wang, Rausch, Zhang, Jia, and Song]{wang_principled_2020}
T.~Wang, J.~Rausch, C.~Zhang, R.~Jia, and D.~Song.
\newblock A {Principled} {Approach} to {Data} {Valuation} for {Federated} {Learning}, Sept. 2020.
\newblock URL \url{http://arxiv.org/abs/2009.06192}.
\newblock arXiv:2009.06192 [cs, stat].

\bibitem[Wu et~al.(2023)Wu, Jia, Lin, Huang, and Chang]{wu_variance_2023}
M.~Wu, R.~Jia, C.~Lin, W.~Huang, and X.~Chang.
\newblock Variance reduced {Shapley} value estimation for trustworthy data valuation, May 2023.
\newblock URL \url{http://arxiv.org/abs/2210.16835}.

\bibitem[Yan and Procaccia(2021)]{yan_core_2021}
T.~Yan and A.~D. Procaccia.
\newblock If {You} {Like} {Shapley} {Then} {You}’ll {Love} the {Core}.
\newblock \emph{Proceedings of the AAAI Conference on Artificial Intelligence}, 35\penalty0 (6), May 2021.
\newblock ISSN 2374-3468.
\newblock \doi{10.1609/aaai.v35i6.16721}.
\newblock URL \url{https://ojs.aaai.org/index.php/AAAI/article/view/16721}.

\bibitem[Zhang et~al.(2024)Zhang, Bi, Cheng, Liu, Ren, Sun, Wu, Cao, Fernandez, Xu, Jia, Kwon, Pei, Wang, Xia, Xiong, Yu, and Zou]{zhang_survey_2024}
J.~Zhang, Y.~Bi, M.~Cheng, J.~Liu, K.~Ren, Q.~Sun, Y.~Wu, Y.~Cao, R.~C. Fernandez, H.~Xu, R.~Jia, Y.~Kwon, J.~Pei, J.~T. Wang, H.~Xia, L.~Xiong, X.~Yu, and J.~Zou.
\newblock A {Survey} on {Data} {Markets}, Nov. 2024.
\newblock URL \url{http://arxiv.org/abs/2411.07267}.
\newblock arXiv:2411.07267 [cs].

\bibitem[Zhu et~al.(2019)Zhu, Dong, Shen, and Gai]{zhu_incentive_2019}
L.~Zhu, H.~Dong, M.~Shen, and K.~Gai.
\newblock An {Incentive} {Mechanism} {Using} {Shapley} {Value} for {Blockchain}-{Based} {Medical} {Data} {Sharing}.
\newblock In \emph{2019 {IEEE} 5th {Intl} {Conference} on {Big} {Data} {Security} on {Cloud} ({BigDataSecurity}), {IEEE} {Intl} {Conference} on {High} {Performance} and {Smart} {Computing}, ({HPSC}) and {IEEE} {Intl} {Conference} on {Intelligent} {Data} and {Security} ({IDS})}, pages 113--118, May 2019.
\newblock \doi{10.1109/BigDataSecurity-HPSC-IDS.2019.00030}.

\end{thebibliography}
